\documentclass[pmlr,twocolumn,10pt]{jmlr} % W&CP article

\usepackage{booktabs}
\usepackage{siunitx}

\usepackage{algorithm}
\usepackage{algorithmic}
\usepackage{array}
\usepackage{booktabs}
\usepackage{multirow}
\usepackage{color}
\usepackage{amsmath}
\usepackage{enumitem}
\usepackage{pifont}
\usepackage{amssymb}
\usepackage{bbding}
\newcommand{\cmark}{\ding{51}}% check mark
\newcommand{\xmark}{\ding{55}}% cross mark
\usepackage[capitalize]{cleveref}
\crefname{section}{Sec.}{Secs.}
\Crefname{section}{Section}{Sections}
\Crefname{table}{Table}{Tables}
\crefname{table}{Tab.}{Tabs.}
\newcommand\DATASETNAME{Radiography Captions}
\newcommand\DATASETNAMEabbr{RGC}

\RequirePackage{xspace}
\makeatletter
\DeclareRobustCommand\onedot{\futurelet\@let@token\@onedot}
\def\@onedot{\ifx\@let@token.\else.\null\fi\xspace}
\def\eg{\emph{e.g}\onedot} 
\def\ie{\emph{i.e}\onedot}

\makeatother

\theorembodyfont{\upshape}
\theoremheaderfont{\scshape}
\theorempostheader{:}
\theoremsep{\newline}

\jmlrvolume{LEAVE UNSET}
\jmlryear{2023}
\jmlrsubmitted{LEAVE UNSET}
\jmlrpublished{LEAVE UNSET}
\jmlrworkshop{Conference on Health, Inference, and Learning (CHIL) 2023} % W&CP title

\title[Multi-modal Pre-training for Medical Vision-language Understanding and Generation]{Multi-modal Pre-training for Medical Vision-language Understanding and Generation: An Empirical Study with \\A New Benchmark}

\author{%
\Name{Li Xu}\Email{li-control.xu@connect.polyu.hk}\\
\Name{Bo Liu} \Email{bokelvin.liu@connect.polyu.hk}\\
\Name{Ameer Hamza Khan} \Email{ameer-hamz.khan@polyu.edu.hk}\\
\Name{Lu Fan} \Email{cslfan@comp.polyu.edu.hk}\\
\Name{Xiao-Ming Wu}\textsuperscript{\Envelope} \Email{xiao-ming.wu@polyu.edu.hk}\\
\addr 
Department of Computing, The Hong Kong Polytechnic University
}

\begin{document}

\maketitle

\begin{abstract}
With the availability of large-scale, comprehensive, and general-purpose vision-language (VL) datasets such as MSCOCO, vision-language pre-training (VLP) has become an active area of research and proven to be effective for various VL tasks such as visual-question answering.
However, studies on VLP in the medical domain have so far been scanty. To provide a comprehensive perspective on VLP for medical VL tasks, we conduct a thorough experimental analysis to study key factors that may affect the performance of VLP with a unified vision-language Transformer.
%We empirically compare different medical VL datasets, visual backbones, and pre-training objectives.
% Specifically, we focus on types of visual backbones, amount and modality of pre-training data, and pre-training objectives.
%Since current medical VL datasets are either noisy or of single modality, we propose \DATASETNAMEINBOLD{}, a multi-modality radiographic dataset containing 18,434 image-caption pairs, collected from an open-access online database MedPix. 
To allow making sound and quick pre-training decisions, we propose RadioGraphy Captions (RGC), a high-quality, multi-modality radiographic dataset containing 18,434 image-caption pairs collected from an open-access online database MedPix. RGC can be used as a pre-training dataset or a new benchmark for medical report generation and medical image-text retrieval.
By utilizing RGC and other available datasets for pre-training, we develop several key insights that can guide future medical VLP research and new strong baselines for various medical VL tasks. 
%Based on the empirical analysis, we develop several key insights which can guide future medical VLP research. Our experiments with RGC demonstrate that a domain-specific dataset with limited high-quality samples is effective for medical VLP. 
\end{abstract}

\paragraph{Data and Code Availability}
In this study, we conduct experiments with 6 public datasets: ROCO~\citep{pelka2018roco}, MedICaT~\citep{subramanian2020medicat}, MIMI-CXR~\citep{mimic-cxr}, SLAKE~\citep{slake}, VQA-RAD~\citep{vqa-rad} and IU X-Ray~\citep{iuxray}. We also collect and filter image-caption pairs from MedPix\footnote{\url{https://medpix.nlm.nih.gov/}} -- an online open-access database and construct the proposed RGC dataset, which is hosted on NIH website\footnote{\url{https://openi.nlm.nih.gov/imgs/collections/RGC.zip}} under the MedPix license. The source code and pre-trained models for reproducing the reported results are available at this link\footnote{\url{https://github.com/Control-xl/Medical-Vision-Langauge-Transformer}}. 
%Before it, we will release the dataset privately to the reviewers in the OpenReview.
% This initial paragraph is \textbf{mandatory}. Briefly state what data you
% use (including citations if appropriate) and whether the data are
% available to other researchers.
% \footnote{An example data availability
% statement: This paper uses the MIMIC-III dataset
% \citep{johnson2016mimic}, which is available on the PhysioNet repository
% \citep{johnson2016physionet}.}
% If you are not sharing code, you must explicitly state that you are not
% making your code available. If you are making your code available, then
% at the time of submission for review, please include your code as
% supplemental material or as a code repository link; in either case, your
% code must be anonymized. If your paper is accepted, then you should
% de-anonymize your code for the camera-ready version of the paper. \emph{If
% you do not include this data and code availability statement for your
% paper, or you provide code that is not anonymized at the time of
% submission, then your paper will be desk-rejected.} Your experiments later
% could refer to this initial data and code availability statement if it is
% helpful (e.g., to avoid restating what data you use).

\paragraph*{Institutional Review Board (IRB)}
{This study has no human-subject research and only uses publicly available and de-identified data, which does not need an IRB approval.}

% This initial paragraph is \textbf{mandatory}. If your research requires IRB
% approval or has been designated by your IRB as Not Human Subject
% Research, then for the camera-ready version of the paper, you must
% provide IRB information (and at the time of submission for review, you
% can say that this IRB information will be provided if the paper is
% accepted). If your research does not require IRB approval, then you
% must state this to be the case. 

\section{Introduction}
\label{sec:intro}
\begin{figure*}[th]
  \centering
  \includegraphics[width=1\linewidth]{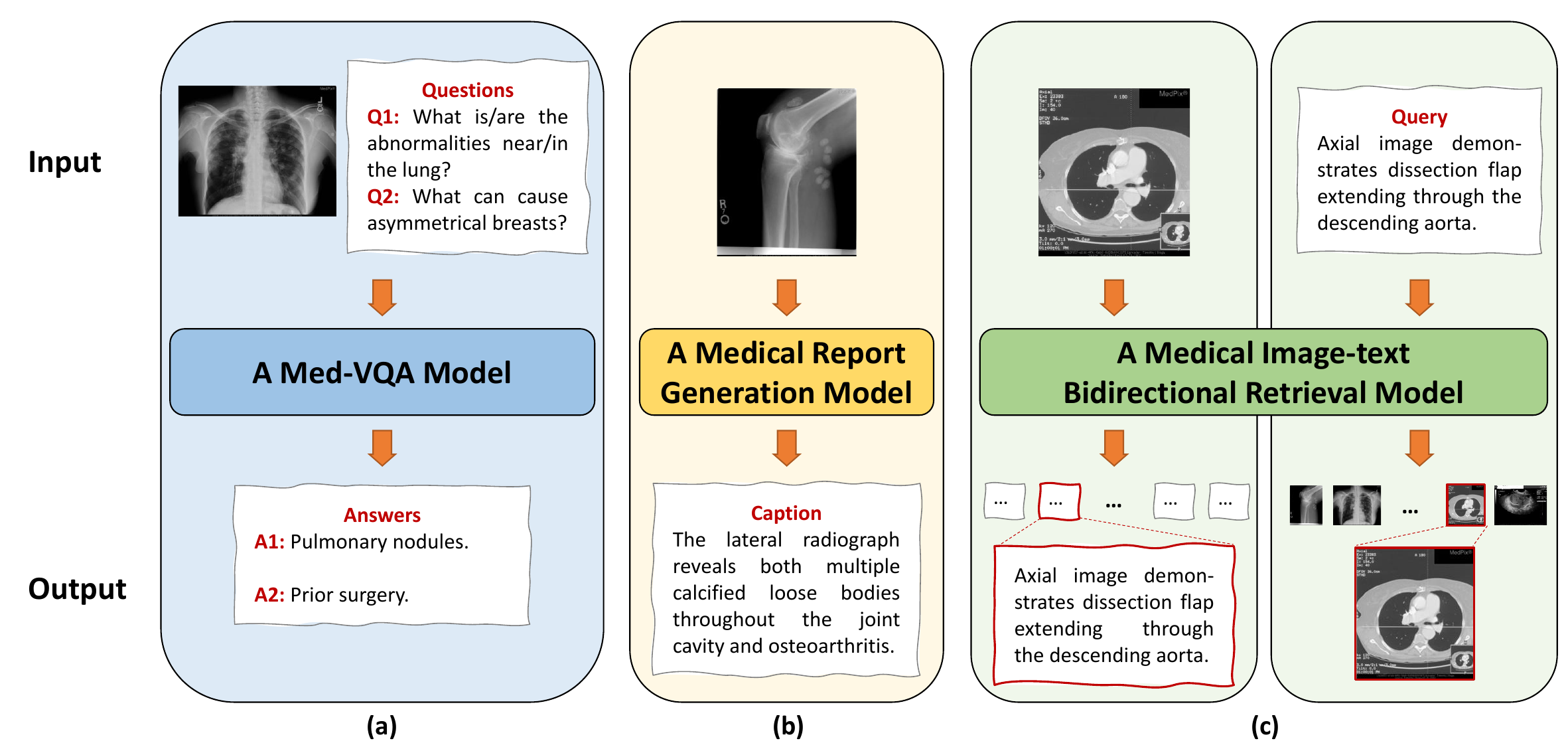}
  \caption{Medical vision-language tasks. (a) Medical visual question answering (Med-VQA). (b) Medical report generation. (c) Medical image-text %bidirectional 
  retrieval.
  }
  \label{fig:intro}
\end{figure*}

\textbf{Background and Motivation.} Medical vision-language (Med-VL) tasks include visual question answering (Med-VQA)~\citep{liu2021contrastive, nguyen2019overcoming}, medical report generation~\citep{r2gen}, and medical image-text retrieval~\citep{zhang2020contrastive}, as illustrated in \Cref{fig:intro}. Due to their great potential in computer assisted diagnosis and healthcare automation, Med-VL tasks have recently attracted increasing attention from the academia. Solving VL tasks requires cross-modal understanding and generation, and vision-language pre-training (VLP) has shown great promise in various VL tasks. 
%have drawn great attention in recent years because of its wide and promising applications in real world. 
%\Cref{fig:intro} shows some potential usage. 
%Report generation models~\citep{r2gen} can greatly expedite the automation of workflows and improve the quality and standardization of health care. % [from R2GEN]
%The development of medical image-text retrieval~\citep{zhang2020contrastive} can help clinician find previous cases and be used to facilitate the landing of AI methods in healthcare industry.  And with a robust medical visual question answering (Med-VQA)~\citep{liu2021contrastive, nguyen2019overcoming} model, patients can better understand their health situations.
%has been proved an effective way to train VL models.
However, unlike the general domain, where the study of VLP has rapidly advanced~\citep{zhou2020unified, li2019visualbert, tan2019lxmert, li2020oscar, kim2021vilt, su2020vlbert, lu2019vilbert,  li2021unsupervised,li2020unimo, gan2020largescale, ALBEF, dou2021meter, chen2020uniter, SOHO},
%~\citep{ su2020vlbert, lu2019vilbert, ALBEF, SOHO},
%have been proposed for VL tasks, 
the research of medical VLP has so far been scanty.

% Several works have been proposed to learn better representations for cross-modal data via pre-training. These models are mainly based on Transformer and is pre-trained to learn joint visual-linguistic features on large-scale image-caption datasets such as MSCOCO~\citep{chen2015mscococaption} and Conceptual Captions~\citep{sharma2018conceptual}. 
% %METER~\citep{dou2021meter} performs empirical studies on VLP based on Transformer in the general domain.

\textbf{Related Works.} % Although significant progress has been made for VLP models in the general domain, little empirical researches has been conducted to explore the performances of pre-trained models in the medical domain.
The main reason may be the lack of publicly available, large-scale, and high-quality medical VL datasets for pre-training.
%which severely hinders the development of VL models in the medical domain. 
Due to privacy concerns and copyright issues, the acquisition of medical image-text data is  difficult. Furthermore, annotating medical data requires significant domain knowledge and medical expertise, which is cost-prohibitive and time-consuming.
%To our knowledge, MMBERT~\citep{khare2021mmbert}, MedViLL~\citep{moon2021multimodal}, Clinical-BERT~\citep{yan2022clinical}, \textcolor{blue}{ARL~\citep{MM2022-VLP} and M3AE~\citep{chen2022m3ae}} are the only works investigating medical VLP. 
So far, there are only a few studies investigating medical VLP, and they are not comprehensive enough. MMBERT~\citep{khare2021mmbert} only targets for one downstream task -- Med-VQA. MedViLL~\citep{moon2021multimodal} and Clinical-BERT~\citep{yan2022clinical} utilize MIMIC-CXR~\citep{mimic-cxr} -- a large single-modality dataset with chest X-ray images for pre-training, and hence the pre-trained models cannot effectively deal with downstream VL tasks involving other imaging modalities or body regions. The most recent works ARL~\citep{MM2022-VLP} and M3AE~\citep{chen2022m3ae} utilize stronger backbone networks
%ROCO and MedICaT for pre-training. ARL introduces 
including pre-trained CLIP-ViT~\citep{radford2021learning} and RoBERTa~\citep{roberta} along with external knowledge or masked autoencoder~\citep{MAE} to improve pre-training performance. However, they cannot deal with generation tasks due to the use of bidirectional Transformer like RoBERTa as text encoder.
%M3AE further utilizes MAE~\citep{MAE} to improve the model performance. However, both of them cannot deal with generation tasks due to their model architectures, which are important tasks in the medical domain.

%ARL~\citep{MM2022-VLP} and M3AE~\citep{chen2022m3ae} are the most similar works to ours, both of which utilize ROCO and MedICaT for pre-training. ARL introduces external knowledge to help the inference. M3AE further utilizes MAE~\citep{MAE} to improve the model performance. However, both of them cannot deal with generation tasks due to their model architectures, which are important tasks in the medical domain.

\textbf{Present Work.} 
To provide a comprehensive perspective on medical VLP,
%we conduct a detailed experimental analysis of VLP in the medical domain based on the general VL Transformer network. 
we conduct a thorough empirical study based on a \emph{unified} framework -- vision-language Transformer (VLT)~\citep{tan2019lxmert, zhou2020unified}, which can deal with \emph{both generation and understanding tasks}. To prepare datasets for pre-training, we examine existing large-scale radiographic VL datasets including  ROCO~\citep{pelka2018roco}, MedICaT~\citep{subramanian2020medicat}, and MIMIC-CXR~\citep{mimic-cxr}.
%Specifically, we compare current medical VL datasets, \ie, MIMIC-CXR\citep{mimic-cxr}, IU X-Ray~\citep{iuxray}, ROCO~\citep{pelka2018roco} and MedICaT~\citep{subramanian2020medicat} when they are individually and jointly used as the pre-training data. 
%
Since these datasets are either noisy or of single imaging modality, to make sound and quick decisions on pre-training settings, we propose to construct a high-quality dataset of diverse radiographic imaging modalities. 
%and helps us make quick decisions on several factors which may affect VLP model capacity. 
Specifically, we collected and filtered image-caption pairs from MedPix
%\footnote{\url{https://medpix.nlm.nih.gov/}}, 
an online open-access database, and manually cleaned both image and text data to obtain 18,434 image-caption pairs, forming the RadioGraphy Captions (RGC) dataset, which can be used for pre-training and making pre-training decisions, or as a benchmark to evaluate the performance of the pre-trained models and existing ones.
%yielding 18,434 image-caption pairs. %  RGC has a higher quality than the original Medpix, which is important in pre-training. As shown in \Cref{fig:medpix_example}, our \DATASETNAMEabbr{} data contains most body parts with diverse imaging modalities. For each radiography image, a caption will be accompanied. 
% Pre-training is quite time-consuming. It may take tens of days to pre-train a VL model on large-scale datasets on a single GPU, unaffordable for researchers with limited computing resources. \DATASETNAMEabbr{} can help alleviate this issue to some extend.
%We further split \DATASETNAMEabbr{} into the training set and testing set, hoping it can be a preliminary benchmark to expand existing medical downstream VL tasks such as report generation and medical image-text retrieval with multiple imaging modalities to meet more actual clinical needs.
With \DATASETNAMEabbr{} and the above mentioned datasets,
we conduct a comprehensive empirical study on pre-training decisions including visual backbone, pre-training objective, and pre-training dataset. Further, we evaluate the effectiveness of the pre-trained VLTs on downstream Med-VL tasks including Med-VQA, report generation, and image-text retrieval, compared with state-of-the-art methods.
%an empirical comparison of visual backbones (\eg, Resnet \citep{resnet}, ViT~\citep{vit} and Swin Transformer \citep{liu2021swin}) and pre-training objectives (\eg, Masked Language Modeling (MLM), Image-text Matching (ITM)). The experiment results provide strong baselines for current medical VL tasks with the novel Transformer-based VLP framework. 
The key findings from our empirical study include:
\begin{itemize}

\item A small but high-quality in-domain dataset is useful for medical VLP and can be more effective than some existing radiographic datasets of much larger size. Data distribution (\eg, diversity of imaging modality), data quantity, and data quality all significantly impact the performance of the pre-trained VLTs.
    %\item A small-scale but high-quality and in-domain dataset with multi-modalities is also effective for pre-training compared with other larger datasets.
    %\item Data quantity, imaging modality, and data quality are all important factors.
    %\item Large-scale pre-training data can lead to steady improvements for the VLP models , but data distribution, modality, and quality should be comprehensively considered.
    \item The pre-trained VLTs demonstrate high effectiveness in downstream understanding tasks including Med-VQA and image-text retrieval. However, they are not effective for medical report generation, which may suggest the inadequacy of the pre-training data and pre-training method for generation tasks. 
    %MLM plays an significant role for understanding tasks. However, with current medical VL datasets, MLM doesn't necessarily contribute to the model performance on generation tasks.
    %\item MLM plays an significant role for understanding tasks. However, with current medical VL datasets, MLM doesn't necessarily contribute to the model performance on generation tasks.
    
    % \item In general, Swin Transformer performs better than other compared visual backbones.
    % \item MLM doesn't improve performances of Transformer encoders on the current medical generation datasets.
\end{itemize}

%\textbf{Contributions.} 
% The main contributions of this work are also two-fold: 
% \begin{itemize}
% \item We present a comprehensive empirical study on medical VLP, providing analysis on pre-training decisions and evaluating the effectiveness of the pre-trained VLTs on three downstream VL tasks. The pre-trained VLTs can also serve as strong baselines for future research.

% \item We propose \DATASETNAMEabbr{}, a high-quality radiographic VL dataset of multiple imaging modalities, which can be used as a pre-training dataset, or as a new benchmark for medical report generation and medical image-text retrieval, to supplement existing single-modality benchmarks.

%\end{itemize}

\section{Unified Vision-language Pre-training} \label{sec:approach}
This section presents a unified vision-language pre-training framework for Med-VL applications including both understanding and generation tasks. 
%Med-VQA, medical report generation, and medical image-text retrieval.

\subsection{Model: Vision-language Transformer} \label{sec:vlt}

\begin{figure*}[t]
  \centering
  \includegraphics[width=1\linewidth]{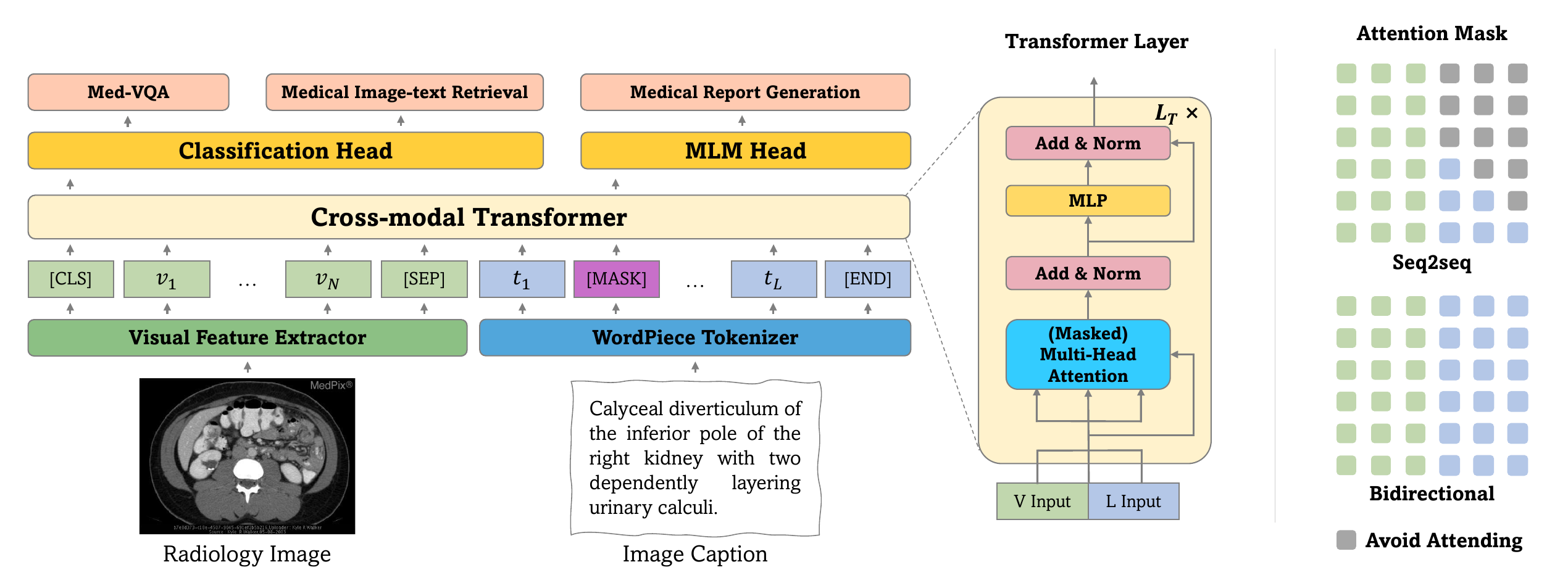}
  \caption{Overall architecture of the unified vision-language Transformer for both pre-training and fine-tuning.
  Both stages share the same network architecture. The main differences are in the input data and the training objectives. For pre-training, the input data is image-caption pairs from the pre-training corpus, with MLM or ITM as the training objective. For fine-tuning, the input data is from the training set of downstream tasks, and the training objective is task-dependent. 
  %Note that our RGC can be used for both pre-training and fine-tuning.
  }
  
  %A visual feature extractor receives medical images to extract feature maps. These feature maps, treated as visual tokens, with corresponding text tokens, which WordPiece generates, are fed into the cross-modal Transformer encoder. 
  % By controlling the attention mask $M$ in the Transformer encoder, we can let the texts only 'see' the text tokens on their left and all the visual tokens along with two special tokens, [CLS] and [SEP].  
  \label{fig:mvlt}
\end{figure*}

%\subsection{Vision-Language Transformer} \label{sec:vlt}
We choose vision-language Transformer (VLT) as the model for pre-training, which has been the model of choice for multi-modal pre-training in the general domain. As shown in \Cref{fig:mvlt} (left), it consists of a visual module, a text module, and a cross-modal Transformer encoder.
% \noindent \textbf{Overview.}
% There are several VLP models based on Transformer (or BERT) in the general domain. 
% Our vision-language Transformer (VLT) is shown in \Cref{fig:mvlt} (left). A typical VLT consists of a visual module, a tokenizer with an optional text encoder, and a multi-modal Transformer encoder composed of several multi-head self-attention layers. %, each followed by a multi-layer perceptron (MLP) and LayerNorm layers. 
%This paper mainly considers three downstream tasks: Med-VQA, report generation, and image-text retrieval. Hence, we should finish understanding, generation, and retrieval tasks in a single pre-training framework. 

\textbf{Visual Module.} 
The visual module is a visual feature extractor that processes the input image and outputs its feature representation $V=[\mathbf{v}_1, \mathbf{v}_2,..., \mathbf{v}_N] \in \mathbb{R}^{N \times d}$, where $N$ is the number of visual tokens, and $d$ is the number of hidden states of the cross-modal Transformer. 
The visual feature extractor can be a pre-trained ResNet as used in some medical vision-language tasks \citep{gong2021cross, r2gen}, or others such as linear patch \citep{kim2021vilt}. Here, we choose Vision Transformer~\citep{vit} and its variant (\eg, Swin Transformer~\citep{liu2021swin}), which have shown promising performance in the general domain but never been used in medical vision-language tasks. Specifically, an image is split into non-overlapping patches, which are treated as visual tokens and fed to the Vision Transformer to extract visual features. 
Note that in the general domain, it is common to treat bottom-up attention~\citep{anderson2018bottom} obtained by an object detector (\eg, Faster R-CNN) as visual tokens for VLT. However, it is not applicable in the medical domain due to the lack of annotated medical images with object labels to train the object detector. 
%In addition, the current object detectors are not perfect.
%pre-trained object detection models such as Faster R-CNN are widely used because treating bottom-up attention \citep{anderson2018bottom} provided by Faster R-CNN as visual tokens are more natural and intuitive for VLT. 
%However, there is no such dataset in the medical domain that aims to train an object detector for all medical images. In addition, the current object detectors are not perfect. Thus, A pre-trained Resnet is preferred \citep{gong2021cross, r2gen}. % khare2021mmbert,
%Recently, Vision Transformer is becoming a trending area, showing the powerful ability of the self-attention mechanism for extracting visual features. Specifically, an image is split into non-overlapping patches, which are treated as visual tokens and further fed into Vision Transformer modules to extract visual features. 
%The study of Vision Transformer in the medical VL domain has so far been scanty. In this paper, we leverage the original Vision Transformer and Swin Transformer, comparing them with Resnet and linear patch \citep{kim2021vilt} (denoted as Linear/P where P is the patch size). 

\textbf{Text Module.}
The text module is a tokenizer for processing the input text sequence. Specifically, we use WordPiece~\citep{wordpiece} to tokenize the text sequence and obtain learnable embeddings $T=[\mathbf{t}_1, \mathbf{t}_2, ..., \mathbf{t}_L]\in \mathbb{R}^{L\times d}$ by embedding lookup, where $L$ is the length of the tokenized sequence. 
Three special tokens, [CLS], [SEP], and [END] with learnable embeddings $t_{CLS}$, $t_{SEP}$, and $t_{END}$ respectively, are also introduced as in many other VL tasks, and so are learnable positional embeddings and segmentation embeddings. Note that similar to previous VLP works \citep{zhou2020unified, yan2022clinical, li2020oscar}, we do not use an additional text encoder such as BERT~\citep{devlin2018bert} as in \citet{dou2021meter} to improve model capacity,  because a bidirectional text encoder is not compatible with the downstream generation task.

%\label{sec:attention_mask}
%\noindent \textbf{Masked Multi-head Self-attention.}

\textbf{Cross-modal Transformer Encoder.} The cross-modal Transformer encoder consists of several multi-head self-attention layers. In each layer, an attention mask is used to control whether one token can attend to others. %so that it can be used for generation tasks. 
%(the graphical illustration of the Transformer Layer and attention mask is inspired by \citep{transformer} and \citep{zhou2020unified})
As illustrated in \Cref{fig:mvlt} (right), a bidirectional attention mask (as used in BERT or RoBERTa ) allows each token to see the tokens on its both sides, whereas a seq2seq attention mask only allows each token to attend to those on its left -- so it can be used for generation tasks. Following Unified VLP~\citep{zhou2020unified}, we use both types of attention masks in the Transformer encoder so the VLP can deal with both understanding and generation tasks. %\textcolor{blue}{We provide a brief discussion on how we fine-tune the model onto downstream task in ~\Cref{sec:fine_tune_downstream}}

We describe how to fine-tune the pre-trained VLT on downstream Med-VL tasks including Med VQA, report generation, and image-text retrieval in \Cref{sec:fine_tune_downstream}.

\subsection{Pre-training Objectives}
To train the VLT, we adopt two widely-used self-supervised pre-training objectives: masked language modeling and image-text matching. %Another possible pre-training objective is Masked Image Modeling (MIM) \citep{dou2021meter, tan2019lxmert}. However, this task doesn't contribute to model capacity according to \citep{dou2021meter}.

\textbf{Masked Language Modeling (MLM).} % MLM is one of the most widely used pre-training objectives in NLP and VL.
MLM~\citep{devlin2018bert} replaces 15\% of text tokens with a special token [MASK], a random token, or the original token with probabilities of 80\%, 10\%, and 10\% respectively.
The model is trained to predict the masked tokens given unmasked ones.
%the original tokens according to the given context.
% As shown in \Cref{fig:mvlt} (left), $t_2$ is masked by replacing it with the special token [MASK]. The output of the Transformer  is used to predict the masked token.
Following Unified VLP~\citep{zhou2020unified}, we train with two sub-objectives, seq2seq MLM and bidirectional MLM, 
%based on what context the tokens can see, which is controlled by the attention mask 
based on different attention masks as introduced in \Cref{fig:mvlt}, which alternates with probability $\alpha$ and $1-\alpha$ respectively.
%The attention mask alternates between bidirectional and seq2seq  with probability $\alpha$ and $1-\alpha$ respectively.

\textbf{Image-text Matching (ITM).} ITM~\citep{tan2019lxmert, li2019visualbert} is similar to next sentence prediction \citep{devlin2018bert}. For each image-text pair, the text will be replaced with another text with a probability of 50\%. The model is trained to determine whether a given image-text pair is matched or not. 
%However, in our experiments, we find ITM doesn't necessarily contribute to pre-trained model performance and is weaker than MLM. 
% This is consistent with the results from \citep{zhou2020unified}. 
%Hence, we choose not to use ITM by default.

 %Hence, we choose not to conduct experiments with MIM.

\begin{figure*}[!t]
    \centering  
    \subfigure[Med-VQA.]{\label{fig:transformer_vqa}
        \includegraphics[width=0.3\linewidth]{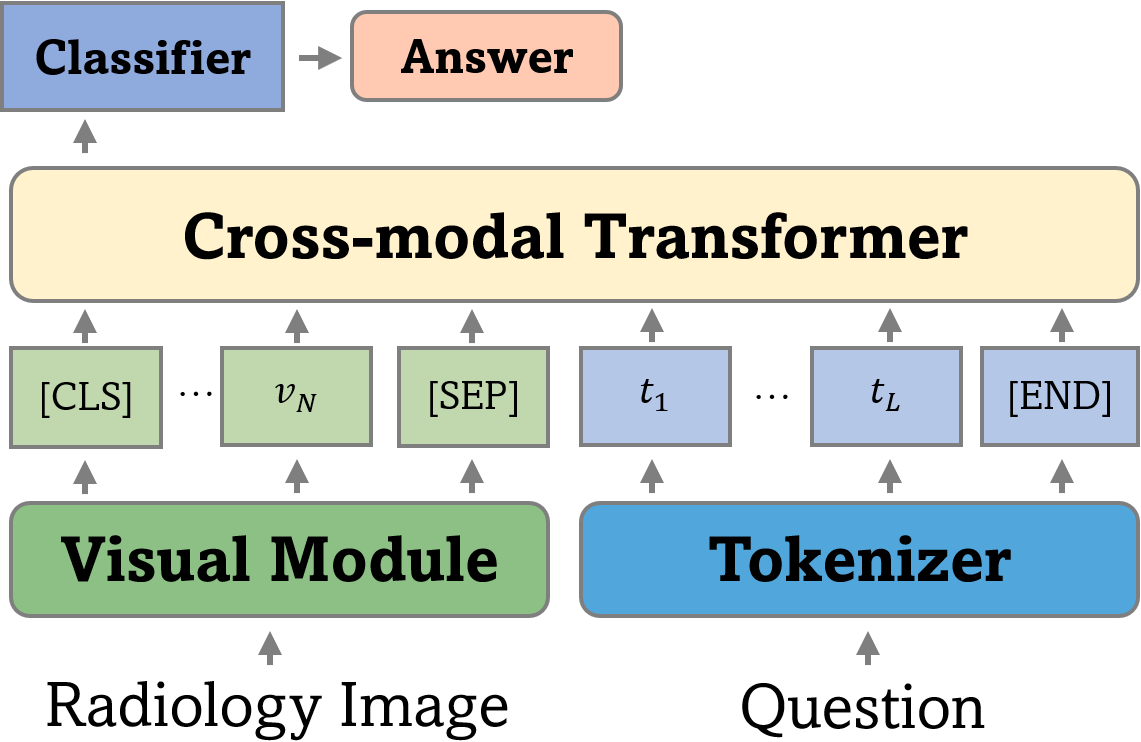}
    }
    \subfigure[Medical report generation.]{\label{fig:transformer_generation}
        \includegraphics[width=0.3\linewidth]{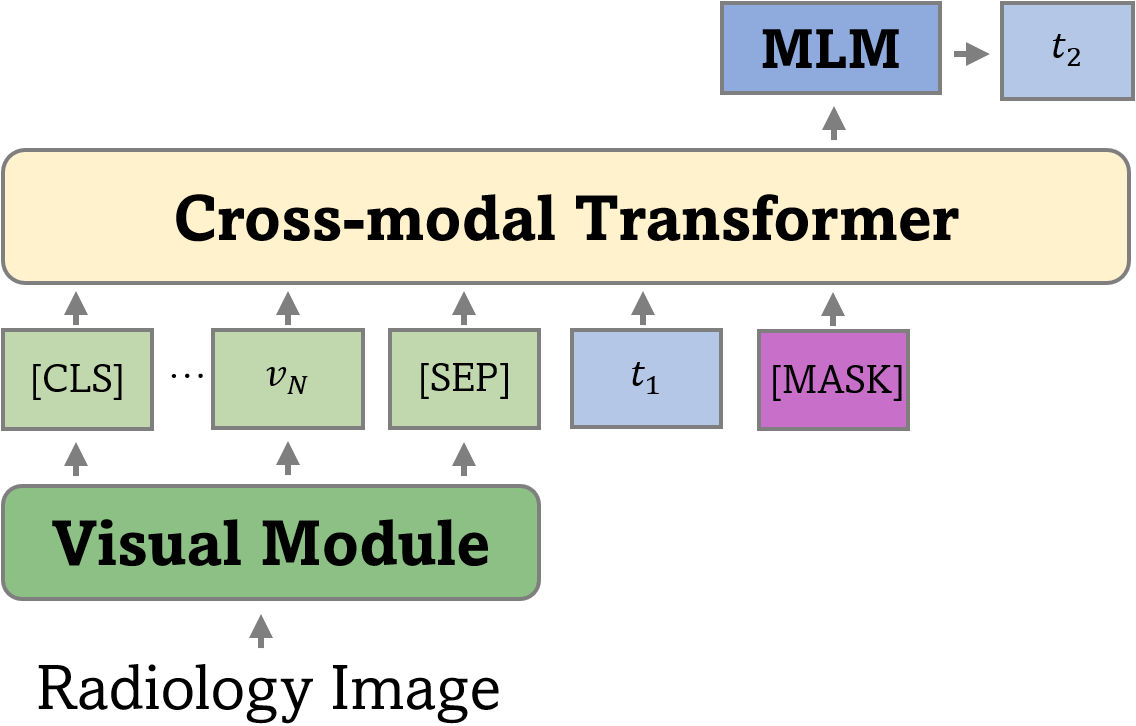}
    }
    \subfigure[Medical image-text retrieval.]{\label{fig:transformer_retrieval}
        \includegraphics[width=0.3\linewidth]{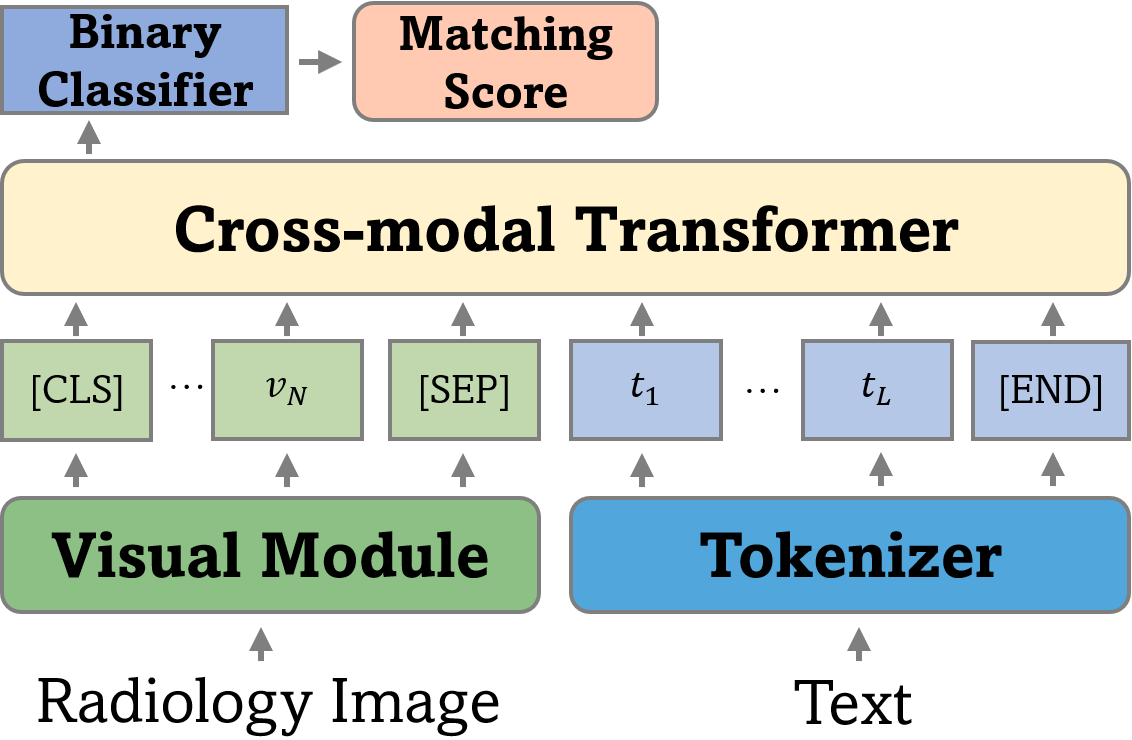}
    }
    % \begin{subfigure}[b]{0.3\linewidth}
    %     \includegraphics[width=\textwidth]{figures/transformer_vqa_small.png}
    %     \caption{Med-VQA.}
    %     \label{fig:transformer_vqa}
    % \end{subfigure}
    % \hfill
    % \begin{subfigure}[b]{0.3\linewidth}
    %     \centering
    %     \includegraphics[width=\textwidth]{figures/transformer_generation_small.png}
    %      \caption{Medical report generation.}
    %      \label{fig:transformer_generation}
    % \end{subfigure}
    % \hfill
    % \begin{subfigure}[b]{0.3\linewidth}
    %     \centering
    %     \includegraphics[width=\textwidth]{figures/transformer_retrieval_small.png}
    %     \caption{Medical image-text retrieval.}
    %     \label{fig:transformer_retrieval}
    % \end{subfigure}
    \caption{Apply the pre-trained and fine-tuned VLT for inference on downstream medical VL tasks.}
     
\end{figure*}

\subsection{Fine-tuning the Pre-trained VLT for Medical Vision-language Tasks} \label{sec:fine_tune_downstream}
In the following, we describe how to fine-tune the pre-trained VLT on downstream medical VL tasks.

\textbf{Medical Visual Question Answering.} 
Similar to VQA, a Med-VQA model aims to find a correct answer given a clinical question related to a medical image. In existing literature, it is commonly formulated as a classification task, \ie, the model is trained to choose the correct answer from a list of candidate answers. To fine-tune the pre-tained VLT, a classifier (\eg, MLP) is attached on top of the output of the [CLS] token by the cross-modal Transformer to predict the answer. The fine-tuned VLT and trained classifier can then be directly used for inference on test data, as shown in \Cref{fig:transformer_vqa}.

%As shown in \Cref{fig:transformer_vqa}, a classifier (\eg, MLP) is attached on top of the output of [CLS] to predict the answer.

\textbf{Medical Report Generation.} 
Given a radiography image, it aims to generate a clinical text that can accurately describe the image. Following Unified VLP~\citep{zhou2020unified}, we use seq2seq MLM as the objective to fine-tune the pre-trained VLT on the training data of this task. The fine-tuned VLT can be directly used for report generation, as illustrated in \Cref{fig:transformer_generation}, which shows a single step of the generation process with the seq2seq attention mask. The token $t_1$ is generated by the previous step, and a special token [MASK] is appended to $t_1$ to predict the next token, \ie, $t_2$, which is generated by a token classifier (\eg, a seq2seq MLM head).

\textbf{Medical Image-text Retrieval.}
Image-text retrieval considers two sub-tasks, image retrieval given text queries and text retrieval given image queries. 
%They are used to evaluate the ability of the model to find image-text pairs.
Following \citep{qi2020imagebert}, we attach a binary classifier on top of the cross-modal Transformer and fine-tune the pre-trained VLT on a training set of image-text pairs. Note that we augment the training data by randomly swapping images or texts to produce negative pairs. The model is trained to determine whether a given image-text pair is matched. For inference, the fine-tuned VLT and trained classifier is directly applied to predict the matching score of any image-text pair, as shown in 
\Cref{fig:transformer_retrieval}. The matching scores will be used for ranking relevant images or texts w.r.t. the query.

\section{Radiographic  Vision-language Datasets for Pre-training}
\label{sec:dataset}

In this section, we briefly introduce existing radiographic VL datasets 
%for pre-training in our experiments along with their weakness and then elaborate on the process of constructing
and present our \DATASETNAMEabbr{} dataset.

\textbf{Existing Radiographic Vision-language Datasets.}
To date, there are five large public radiographic VL datasets that can be potentially used for VLP, including MIMIC-CXR~\citep{mimic-cxr}, IU X-Ray~\citep{iuxray}, MedICaT~\citep{subramanian2020medicat},  and ROCO~\citep{pelka2018roco}. However, as summarized in \Cref{tab:stat_dataset_pretraining}, they all have limitations. Both MIMIC-CXR and IU X-Ray only include \emph{images of a single modality on a specific body region} -- chest X-ray images. Many images in MedICaT have multiple sub-figures whose captions are concatenated to form a text, resulting in weak image-caption alignment. 
%and it is curated to focus on the alignment between sub-captions and sub-figures.
%which will fail to make themselves effective datasets to evaluate model performances on report generation and image-text retrieval with different image modalities. 
%Current medical VL datasets are MIMIC-CXR~\citep{mimic-cxr}, IU X-Ray~\citep{iuxray}, MedICaT~\citep{subramanian2020medicat} and ROCO~\citep{pelka2018roco}. Both MIMIC-CXR and IU X-ray contain chest X-ray images and MedICaT mainly focuses on the alignment between sub-captions and sub-figures, which will fail to make themselves effective datasets to evaluate model performances on report generation and image-text retrieval with different image modalities. 
%The dataset most similar to \DATASETNAMEabbr{} is ROCO. 
ROCO is built on archives/documents from PubMed Central\footnote{\url{https://www.ncbi.nlm.nih.gov/pmc/}} and automatically filters out non radiographic images and images with sub-figures. However, it may inevitably contain invalid captions or images with unwanted annotations. There are other VL datasets including CheXpert~\citep{irvin2019chexpert}, BIMCV COVID-19+~\citep{vaya2020bimcv}, PadChest~\citep{padchest}, and
FFA-IR~\citep{li2021ffair}. However, CheXpert is not a public dataset; BIMCV COVID-19+ is in Spanish; PadChest is similar to MIMIC-CXR but with a smaller size; FFA-IR contains fundus fluorescein angiography images and is of single modality like MIMIC-CXR. Hence, we do not use them for pre-training. 
%Other possible datasets for VLP include PadChest~\citep{padchest}, CheXpert~\citep{irvin2019chexpert}, BIMCV COVID-19+~\citep{vaya2020bimcv}, and FFA-IR~\citep{li2021ffair}. PadChest and CheXpert also focus on chest X-ray. We argue MIMIC-CXR can be a representative for them.
%FFA-IR mainly focus on fundus fluorescein angiography, which is also of single modality. The report of BIMCV COVID-19+ are in Spanish.

%, which can lower the data quality. 
\begin{figure}[t!]
    % \centering  }
        % \includegraphics[width=\textwidth]{figures/rgc/adb-ct.png}
    % \includegraphics[width=0.45\textwidth]{figures/medpix_example4.png}
    \includegraphics[width=0.5\textwidth]{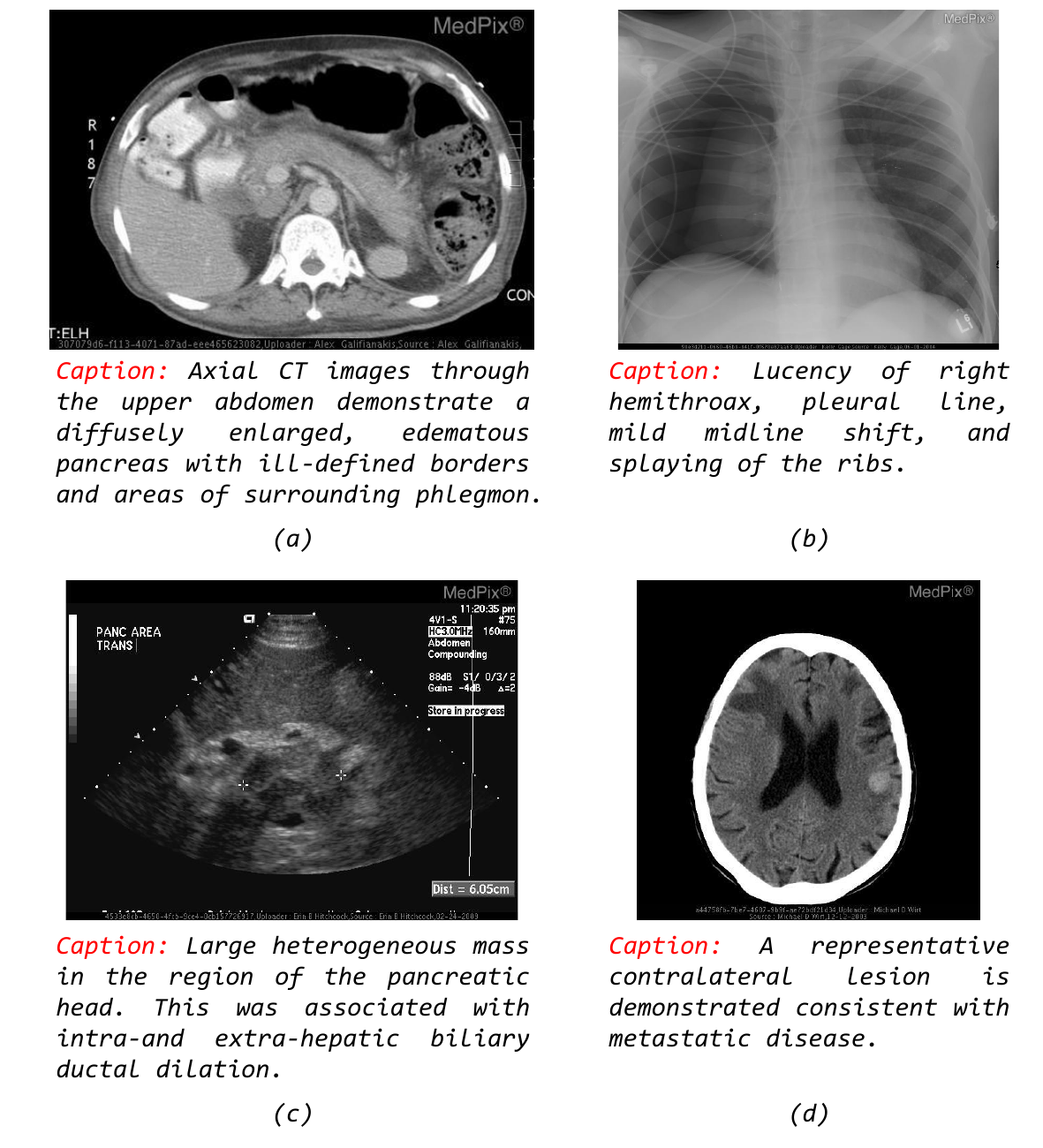}
    \caption{Samples from our \DATASETNAMEabbr{} dataset with various radiographic imaging modalities. (a) Abdomen CT. (b) Chest X-ray. (c) Liver ultrasound. (d) Brain MRI.}
     \label{fig:medpix_example}
\end{figure}

\begin{table*}[t!]
    \centering
    \scalebox{1}{
    \begin{tabular}{c|ccccc}
    \toprule
     Modality      & MRI & CT  &  X-ray & Ultrasound & Not provided  \\
    \midrule
    \# samples & 6471 & 6922 & 3510 &  832 &  699 \\

    \bottomrule
    \end{tabular}
    }
    \caption{The statistics of imaging modalities of \DATASETNAMEabbr{}.}
    \label{tab:rgc_modality}
\end{table*}

\begin{table*}[t!]
    \centering
    \scalebox{1}{
        \begin{tabular}{c|ccccc}
        \toprule
        Datasets         & ROCO   & MedICaT
        & MIMIC-CXR   & IU X-Ray % & FFA-IR
        %& MedPix 
        & \DATASETNAMEabbr{} \\
        \midrule
        \# image-caption pairs           & 87952  & 217060 & 473057     & 7470     
        & 18434 \\
        Multiple imaging modalities      & \cmark & \cmark  & \xmark     & \xmark  % & \xmark 
        & \cmark\\
         Various body regions           & \cmark & \cmark  & \xmark     & \xmark  %& \xmark 
        & \cmark\\
         Good image-caption alignment    & \cmark & \xmark  & \cmark  % & \cmark  
        & \cmark  & \cmark  \\
         Manually cleaned          & \xmark & \xmark & \cmark & \xmark  %& \cmark
         &  \cmark \\
       Rich annotations  & \xmark & \xmark & \xmark & \xmark  % & \cmark 
        &   \cmark \\
        % & \cmark\\
        \bottomrule 
        \end{tabular}
    }
    \caption{Comparison of different radiographic  vision-language datasets.}
    \label{tab:stat_dataset_pretraining}
\end{table*}

We propose to construct a high-quality radiographic VL dataset with various imaging modalities, which can be effectively used as a pre-training corpus and for making pre-training decisions, as well as a medical VL benchmark for evaluating the performance of the pre-trained models and existing methods.
%as part of medical VLP and as a benchmark for downstream medical VL tasks to evaluate the performance of the pre-trained models as well as existing VL methods.

\subsection{Construction of RGC Dataset}
%Our goal is to construct a high-quality multi-modal medical dataset consisting of pairs of radiography images with associated captions, which is intended to be effectively used for medical VLP as well as a benchmark for downstream medical VL tasks.
The construction of our proposed \DATASETNAME{} (RGC) dataset is built on MedPix, a free open-access online database with 37,997 image-caption pairs\footnote{As of the submission of this manuscript, the number has grown to over 59,000, but the newly added cases are not downloadable yet. %Hence, we may continue to expand \DATASETNAMEabbr{} in the future.
}. 
MedPix is intended to train medical students and physicians. %, where each case includes an image, a summary caption, a detailed explanation note, and other useful case-related information. 
Each case in MedPix contains multiple image-caption pairs of different imaging modalities (e.g., MRI, CT and Ultrasound), as well as detailed descriptions and rich annotations, including history, findings, diagnosis, treatment, and discussion, providing valuable resources for medical AI research. 
%For example, the diagnosis label can be used to train a disease classification model; the treatment label can be used to train a treatment plan generation model; the disease and treatment information can be used to build a knowledge base for recommendation or question-answering systems; the multiple image-caption pairs can be used to train a multi-view report generation model.  In contrast, each case in ROCO and MedICaT only contains a single image-caption pair without additional annotations.
For each case, we form image-caption pairs using all images under the case and the summary caption associated with each image.

%Hence, MedPix is a valuable resource for training and evaluating medical VL models.
%In this paper, we mainly focus on the images and their captions.
However, MedPix also includes some incomplete cases. Since captions are not compulsory, some of them are placeholders or invalid, and useful information may be provided in the explanation note but not in the caption.
%, which may affect training of the model. 
In addition, some images and captions contain noisy information. Hence, it is necessary to clean the raw data.
%Thus, we propose to clean both caption and image data to achieve our goal. Next, we briefly describe how we clean caption and image data from MedPix. Note that we clean the data under the hypothesis that if one image is paired with a meaningful caption, we assume that the caption correctly describes the image.

\textbf{Filtering Based on Images and Captions.} 
First, roughly 1,500 images are automatically removed because the caption is too short or meaningless. Second, about 10,000 images are manually inspected and removed. Next, about 8,000 images are manually removed because the captions are not useful for training a machine learning model. 
%Specifically, in this process, we remove non-radiology images, images with too many sub-figures, images with teaching annotations (e.g., arrows, words), and images whose caption contains specific numerical information (e.g., ``a 4 mm osteochondral defect'') or comparative descriptions (e.g., ``compared to prior diastolic image'') involving multiple images.}
Specifically, the manual cleaning process includes

\begin{itemize}
    \item Manually removing non-radiology images or images with too many (\eg, $>16$) sub-figures.  
    %, we will discard it. But for images with low resolution, we decrease the threshold to 9.
   
    \item Manually removing images with teaching annotations (e.g., arrows, words), because these marks may introduce bias during training. 
    \item Manually removing images whose caption contains specific numerical information (e.g., ``tumor with 2 cm area of central necrosis'', ``a 4 mm osteochondral defect'', ``a previous CT 4 days prior'') or comparative descriptions (e.g., ``compared to prior diastolic image'') that involve multiple images, because these information cannot be learned by existing machine learning models,  and including them would lower the quality of RGC as a benchmark.
\end{itemize}

\begin{table*}[t]
     \centering
    \scalebox{1}{
    \begin{tabular}{l|c|c|c|c|c}
    \toprule
     Body region & Head & Chest &   Abdomen & Neck & Pelvic cavity \\
    \midrule
     VQA-RAD   & 104(CT/MRI)& 107(X-Ray) &  104(CT)  & - & -\\
      SLAKE    & 140(CT/MRI)&  219(X-ray)  & 201(CT/MRI) & 41(CT) & 41(CT)\\
    \bottomrule
    \end{tabular}
    }
    \caption{Statistics of Med-VQA datasets.} % Note the numbers for pre-training may differ from the original datasets as we filter out some samples.
\label{tab:stat_med_vqa_data}
\end{table*}

After manual cleaning, we obtain 18,434 image-caption pairs. We then fix typos and clean noisy words from the captions, i.e., redundant words that do not have semantics (e.g., ``Figure 1'' and HTML tags). 
%We further fixed typos and manually removed noisy words (\eg,  ``Figure 1'' and HTML tags) that do not have semantics in the captions to improve data quality.  
%This process is automatic and reproducible. 
%The entire data cleaning process helps to remove noisy images and captions so RGC can be used as a better benchmark for medical vision-language tasks. 
Finally, we divide the 18,434 clean image-caption pairs into a training set and a test set with a ratio of 9:1, which can be used as a new multi-modality benchmark for medical report generation and image-text retrieval. \Cref{tab:rgc_modality} shows the statistics of imaging modalities of \DATASETNAMEabbr{}. Note that most of existing benchmarks such as MIMIC-CXR~\citep{mimic-cxr} and IU X-Ray~\citep{iuxray} are of single modality.

% \begin{itemize}
%     \item \textcolor{blue}{Manually removing non-radiology images or images with too many sub-figures.}
%     \item\textcolor{blue}{Manually removing images with teaching annotations (e.g., arrows, words), because these marks may introduce bias during training. }
%     \item \textcolor{blue}{Manually removing images whose caption contains specific numerical information (e.g., ``tumor with 2 cm area of central necrosis'', ``a 4 mm osteochondral defect'', ``a previous CT 4 days prior'') or comparative descriptions (e.g., ``compared to prior diastolic image'') that involve multiple images, because these information cannot be learned by existing machine learning models,  and including them would lower the quality of RGC as a benchmark.}
% \end{itemize}

% \textcolor{blue}{Finally, we divide the 18,434 clean image-caption pairs into a training set and a test set with a ratio of 9:1, which can be used as a new multi-modality benchmark for medical report generation and image-text retrieval. \Cref{tab:rgc_modality} shows the statistics of imaging modalities of \DATASETNAMEabbr{}. Note that most of existing benchmarks such as MIMIC-CXR~\citep{mimic-cxr} and IU X-Ray~\citep{iuxray} are of single modality.}  

\section{Experiments}
\label{sec:experiment}

In this section, we conduct extensive experiments to study the effectiveness of unified VLP for various medical VL tasks.
%pre-training settings affect the performance of VLT.
We first compare different pre-training objectives, visual modules, and pre-training datasets. Then, we report the results of VLT on three downstream tasks.
We also report additional experimental results in Appendix \ref{sec:appendix}.
%the supplementary material. % and the statistics of datasets for downstream tasks

\subsection{Implementation Details}
% in order to have a fair comparison between Resnet and visual Transformer. All the images are resized into $(3,224,224)$ and the feature map of shape $(2048,7,7)$ after the adaptive pooling layer is used and further transposed into $(49,768)$ by a fully connected layer as the visual input for the VLT.
We conduct experiments on PyTorch\footnote{\url{https://pytorch.org/}}.
The structure of cross-modal Transformer is the same as BERT-base\citep{devlin2018bert} from HuggingFace\footnote{\url{https://huggingface.co/transformers/index.html}} \citep{huggingface}, which has 12 Transformer layers with 12 attention heads and 768 hidden units. 
% For visual features, we explore the performance of two popular backbones, namely Resnet-101 and Swin Transformer, both pre-trained on ImageNet~\citep{deng2009imagenet}. 
% In order to have a fair comparison between Resnet and Swin Transformer, 
As for visual backbones, we conduct experiments on ResNet~\citep{resnet}, Vision Transformer~\citep{vit}, Swin Transformer~\citep{liu2021swin}, and linear patch~\citep{kim2021vilt}.
We use the pre-trained weights of ResNet101 and ViT-B/16 from PyTorch.
We use the official implementation of Swin-S\footnote{\url{https://github.com/microsoft/Swin-Transformer}}.
All the images are resized into $(3,224,224)$. In ResNet and Swin Transformer, the output of the last adaptive pooling layer with size $(2048, 7, 7)$, is transposed into $(49, 768)$ by a fully connected layer and then fed into the cross-modal Transformer. 
During pre-training, the alternating probability $\alpha$ between 2 types of attention masks is set to 0.5 during pre-training. In the fine-tuning stage, $\alpha$ is set to 1 on Med-VQA and image-text retrieval tasks (\ie, the attention mask is always bidirectional), and 0 on report generation tasks.
All experiments reported are conducted on a single NVIDIA GeForce RTX 3090 24GB. The code for the experiments and materializing the dataset is submitted along with the paper in the OpenReview.%The dataset is available at open-i\footnote{\url{https://openi.nlm.nih.gov/imgs/collections/RGC.zip}}. The code is available at GoogleDrive\footnote{\url{https://drive.google.com/file/d/1Grub8lQzghQI7j2thVuVj4HfNzt2-TcL/view}}.

\textbf{Optimizer.} We use AdamW optimizer~\citep{adamw} for all the experiments. The weight decay is set to 1e-4. 
During pre-training, the learning rate is 4e-5.
On the downstream tasks, the learning rate is set to 1e-6 for image-text retrieval, and 2e-5 for Med-VQA and 1e-5 for report generation. 

\textbf{Batch Size.}
The batch size for pre-training and report generation is 32, and 64 for others. When ViT-B is the visual backbone, with a batch size of 64, it will lead to an out-of-memory error for a 24GB GPU and hence we reduce the batch size to 48 on downstream tasks.

%\subsection{Impact of Visual Modules}
\subsection{%Ablation Studies of the 
Influence of Pre-training Settings}\label{sec:pretraining_set}
% \subsection{Impact of Pre-training Objectives} \label{sec:impact_of_obj}

Following the common practice in~\citep{dou2021meter}, we mainly conduct empirical comparisons on Med-VQA tasks, since it is the fastest way to assess the comprehension ability of the pre-trained models. We experiment with two Med-VQA benchmarks: VQA-RAD~\citep{vqa-rad} and SLAKE~\citep{slake}. VQA-RAD contains 315 images also from MedPix with 3064 question-answer pairs for training and 451 for testing. In this paper, we use the English version of SLAKE, which contains 642 radiology images with 4919 question-answer pairs for training, 1053 for validation, and 1061 for testing.
\Cref{tab:stat_med_vqa_data}  shows the statistics of Med-VQA datasets w.r.t body regions.
For SLAKE, there are 282 CT images, 181 MRI images, and 179 X-Ray images. For VQA-RAD, they do not provide detailed modality information.
We report average accuracy (\%) with standard deviation over 10 runs.
%(each with 100 epochs).}
0.8

\textbf{On Pre-training Objectives.} We use the small version of Swin Transformer (Swin-S) as the visual backbone to compare different pre-training objectives. The results are provided in \Cref{tab:pretrain_obj_med_vqa}, where we can draw the main observation:
\begin{itemize}[leftmargin=5.5mm]
\item MLM is much more effective than ITM as a pre-training objective, and the latter is not consistently useful. % have a negative effect on the performance of the pre-trained model.
%limit the model performance in the medical domain.
\end{itemize}
Different from the observation in \citep{dou2021meter}, ITM does not help to improve model performance in medical VLP, which is an echo of the finding in~\citep{zhou2020unified}. 
It is probably due to the differences in image modality and scale of pre-training data. The radiology images tend to be similar, and ITM may not be adequate for the model to learn meaningful representations, whereas in the general domain there is often a large difference between two images. 
In addition, the scale of the pre-training data for general VLP is around 10M, much larger than that used in our pre-training. Since the number of negative samples in our pre-training data is limited, ITM is less effective.

\begin{table}[t]
        \centering
        \scalebox{0.75}{
        \begin{tabular}{c|c|c|c}
        \toprule
        % Datasets & \multirow{2}{5em}{Pre-training Objectives} &  VQA-RAD &  SLAKE \\% \multicolumn{2}{c}{Pre-training Objectives}   \\
        \multirow{2}{*}{Datasets} & \multirow{2}{6em}{Pre-training Objectives} & \multirow{2}{*}{VQA-RAD}  &  \multirow{2}{*}{SLAKE}  \\%
         & & \\ 
        \midrule
        \multicolumn{2}{c|}{No Pre-training} 
          &  65.28 $\pm$ 0.98 &  79.74 $\pm$ 0.97  \\ 
        \midrule
        \multirow{3}*{\DATASETNAMEabbr{}}& MLM      &   \bf69.77 $\boldsymbol\pm$ 0.66 &  \bf81.50 $\boldsymbol\pm$ 0.36\\
        & ITM      &  65.47 $\pm$ 1.01 &  80.31 $\pm$ 0.95\\
        & MLM+ITM  &  67.54 $\pm$ 1.16 &  80.71 $\pm$ 0.45\\
        \midrule
        \multirow{3}*{\DATASETNAMEabbr{}+ROCO}& MLM &  \bf 70.43 $\pm$ 0.51 &  \bf 82.34 $\pm$ 0.44 \\ % \bf69.36 $\boldsymbol{\pm}$ 0.86 &  \bf82.92 $\boldsymbol{\pm}$ 0.38 \\
        & ITM &  64.96 $\pm$ 1.03 &  81.18 $\pm$ 0.70    \\
        & MLM+ITM & 67.29 $\pm$ 1.23 &  82.27 $\pm$ 0.57  \\ 
        \bottomrule
        \end{tabular}
        }
        \caption{Comparison of pre-training objectives with Swin-S as the visual backbone. Average accuracy (\%) with standard deviation over 10 runs is reported.  }
        \label{tab:pretrain_obj_med_vqa}
\end{table}

\textbf{On Visual Backbones.} 
We compare different visual backbones in \Cref{tab:visual_backbone_med_vqa}.
%shows the results on Med-VQA.
All the models with different visual backbone are first pre-trained on \DATASETNAMEabbr{} with MLM and then fine-tuned on Med-VQA datasets.
We can make the following observations: 
\begin{itemize}[leftmargin=5.5mm]
    \item With each different visual backbone, pre-training on \DATASETNAMEabbr{} helps to boost the performance of VLT on both Med-VQA datasets.
    \item Visual backbones with locality (\ie, Resnet and Swin Transformer) outperform those without it, and Swin Transformer achieves the best overall performance on both datasets. 
    %On both datasets, Swin Transformer outperform other visual backbones.
\end{itemize}
% Among these backbones, Swin Transformer get the best average scores. 
% when pre-trained by \DATASETNAMEabbr{}. % Swin-S gets the best average scores on Med-VQA.
We can also observe that Linear/16 achieves comparable performance with Swin-S on VQA-RAD after pre-training, but fail to generalize well on SLAKE. It is probably because the number of images in VQA-RAD is quite limited.

\begin{table}[!t]
       \centering
        \scalebox{0.88}{
        \begin{tabular}{l|c|c}
        \toprule
        Models          &  VQA-RAD &  SLAKE   \\
        \midrule
        Linear/16 (w/o pt)    &  59.87 $\pm$ 1.46 &  73.71 $\pm$ 1.14\\ 
        Linear/16             &  69.56 $\pm$ 0.96 &  78.23 $\pm$ 0.36  \\
        ViT-B/16 (w/o pt)     &  61.97 $\pm$ 1.35 &  75.45 $\pm$ 0.61 \\ 
        ViT-B/16              &  67.81 $\pm$ 1.07 &  79.25 $\pm$ 0.48  \\ 
        Resnet-101 (w/o pt)   &  64.34 $\pm$ 1.48 &  77.58 $\pm$ 0.67  \\ 
        Resnet-101            &  69.62 $\pm$ 1.03 &  80.03 $\pm$ 0.49  \\ 
        Swin-S  (w/o pt)      &  65.28 $\pm$ 0.98 &  79.74 $\pm$ 0.97  \\   
        Swin-S                &  \bf69.77 $\boldsymbol\pm$ 0.66 &  \bf81.50 $\boldsymbol\pm$ 0.36\\
        \bottomrule
        \end{tabular}
        }
         \caption{Comparison of vision modules. Average accuracy (\%) with standard deviation over 10 runs is reported. The models are pre-trained on \DATASETNAMEabbr{}. % Linear: images are processed into linear patches and then fed into VLT;    
        pt: pre-training; B: base; S: small. }
        %P in Visual/P denote the patch size to process images.
        \label{tab:visual_backbone_med_vqa}
\end{table}

\begin{table*}[t!]
    \centering
     \scalebox{0.95}{
        \begin{tabular}{ccccc|c|c|c}
        \toprule
        \multicolumn{5}{c|}{\large Pre-training datasets} &  & \multicolumn{2}{c}{\large Accuracy (pre-trained model)} \\
         \DATASETNAMEabbr{} &  ROCO & MedICaT & MIMIC-CXR &  MedPix& \# samples & VQA-RAD &  SLAKE  \\
        \midrule
        \xmark & \xmark &  \xmark  & \xmark  &  \xmark  &  0            & 65.28 $\pm$ 0.98&  79.74 $\pm$ 0.97  \\

        \cmark  & \xmark  & \xmark  & \xmark  & \xmark & $\sim$17k    & 69.77 $\pm$ 0.66 & 81.50 $\pm$ 0.36\\
        \xmark  & \cmark  & \xmark  & \xmark  & \xmark & $\sim$65k    & 70.31 $\pm$ 0.62 &  82.16 $\pm$ 0.43\\  
        \xmark  & \xmark  &\cmark & \xmark &\xmark &  $\sim$217k   & 68.92 $\pm$ 1.26 &  81.81 $\pm$ 0.65\\ 
        \xmark  & \xmark &\xmark & \cmark & \xmark &  $\sim$271k   & 65.34 $\pm$ 1.01 &  81.53 $\pm$ 0.48\\ 
         \xmark  & \xmark  & \xmark  & \xmark  &  \cmark & $\sim$38k   & 69.53 $\pm$ 0.59 &   81.54 $\pm$ 0.51  \\
        % \midrule
       
        \cmark  & \cmark  & \xmark  & \xmark  & \xmark & $\sim$82k    & 70.43 $\pm$ 0.51 &  82.34 $\pm$ 0.44 \\
        \cmark  & \xmark  & \cmark  & \xmark  &   \xmark & $\sim$234k  & 70.07 $\pm$ 0.84 & 82.18 $\pm$ 0.47  \\
         \xmark  & \cmark  & \cmark  & \xmark  & \xmark & $\sim$282k   &  \bf71.25 $\boldsymbol\pm$ 1.02 &  83.08 $\pm$ 0.55 \\
        
        \cmark  & \cmark  & \cmark  & \xmark  & \xmark & $\sim$299k   & \bf71.20 $\boldsymbol\pm$ 0.83 &  \bf83.40 $\boldsymbol\pm$ 0.57 \\ 
        \cmark  & \cmark  & \cmark  & \cmark  &  \xmark &  $\sim$570k   & 70.09 $\pm$ 0.76 &  83.13 $\pm$ 0.54\\ 
        \bottomrule
        % \multicolumn{8}{l}{} \\
        % \multicolumn{8}{l}{}
        \end{tabular}
    }
    \caption{
    %Performance comparison for VL model (Cross-modal Transformer Encoder with Swin-S backbone) when pre-trained on selected combinations of datasets.
    Comparison of different pre-training datasets and their combinations. 
    Average accuracy (\%) with standard deviation over 10 runs is reported.
    \cmark~ means used in pre-training, and \xmark~ means not.}
    %  Comparison results with selected combinations of pre-training datasets. The increase in pre-training data also increases the accuracy
    \label{tab:data_amount_med_vqa}
\end{table*}

% \subsection{Comparison of Pre-training Data} \label{sec:impact_of_data}
\textbf{On Pre-training Datasets.}
We compare the medical VL datasets introduced in \Cref{tab:stat_dataset_pretraining}. We also try using the unfiltered raw MedPix data (about 38k image-caption pairs) for pre-training.
We use Swin-S as the visual backbone and MLM as the pre-training objective. 
The results are summarized in \Cref{tab:data_amount_med_vqa}. 
Note that the statistics of image-caption pairs (\# of samples) may slightly differ from the original datasets as we filter out some invalid samples. 
According to the results, we can make the following observations:
\begin{itemize}[leftmargin=5.5mm]
    \item The best corpus for pre-training is the combination of \DATASETNAMEabbr{}, ROCO and MedICaT. Further including MIMIC-CXR for pre-training decreases model performance. It shows both data quantity and data distribution are important, and the large mass of single-modality data in MIMIC-CXR may introduce bias during pre-training.

    \item The VLT pre-trained with \DATASETNAMEabbr{} can achieve comparable results with those pre-trained 
    %other models pre-trained by datasets of much
    with much larger datasets (\eg, MedICaT, MIMIC-CXR), demonstrating its effectiveness.
    
    \item The single-modality dataset MIMIC-CXR is the least effective for pre-training, despite being the largest dataset, which shows the importance of pre-training with multi-modality images.
    %The dataset Single modality dataset (\ie, MIMIC-CXR) is less effective than multi-modal datasets.
    
     \item The VLT pre-trained with ROCO outperforms the one pre-trained with \DATASETNAMEabbr{}, showing the influence of dataset scale on pre-training. Note that while the differences in performance are small, the results are statistically significant (the p-values for differences between RGC and ROCO are 0.076 on VQA-RAD and 0.00056 on SLAKE).

    \item The VLT pre-trained with RGC performs slightly better than the one pre-trained with MedPix. Note that the size of RGC is less than half of MedPix, which demonstrates the effectiveness of our data cleaning strategy.
    %RGC is slightly better than MedPix on VQA-RAD, and comparable with MedPix on SLAKE. Note that the size of RGC is less than half of MedPix, which demonstrates the effectiveness of our data cleaning strategy. 
    
    %\item Large-scale datasets for pre-training can lead to steady improvements, but data distribution and quality should be also considered comprehensively. When \DATASETNAMEabbr{}, ROCO and MedICaT are jointly used for pre-training, the model performs best. But the usage of MIMIC-CXR reduces the scores. %, which indicates that the amount of in-domain data plays a significant role in pre-training.
\end{itemize}

\begin{table}[t!]
    \centering
    \scalebox{1}{
    \begin{tabular}{l|c}
    \toprule
    Pre-training Datasets  & VQA-RAD   \\
    \midrule
    RGC w/ overlapping images  &   69.77 $\pm$ 0.66   \\
    RGC w/o overlapping images  &   69.80 $\pm$ 0.84  \\
    \bottomrule
    \end{tabular}
    }
    \caption{
    %Comparison of model performance when pre-trained with or without the overlapping images between RGC and VQA-RAD.
    Evaluating the impact of pre-training with overlapping images from RGC and VQA-RAD on model performance.} % Note the numbers for pre-training may differ from the original datasets as we filter out some samples.
\label{tab:overlapping}
\end{table}

\begin{table*}[t]
    \centering

    \scalebox{0.82}{
    \begin{tabular}{c|ccc|ccc}
    \toprule

    &  \multicolumn{3}{c|}{VQA-RAD}&\multicolumn{3}{c}{SLAKE}\\
     Models   &  Overall &  Open-ended & Closed-ended & Overall &  Open-ended & Closed-ended  \\
    \midrule
    % MFBCoAtt*~\citep{yu2017multimodal} &50.6 &14.5 &74.3 &73.3 &72.2 &75.0 \\
    SAN ~\citep{vqa-rad,yang2016stacked} &54.3 & 31.3& 69.5& 76.0&74.0 &79.1 \\
    MFH~\citep{yu2018beyond} &57.9 & 35.2& 72.8&75.9 &73.6 &79.3 \\
    MCB~\citep{vqa-rad,fukui2016multimodal}& 58.1&38.0 &71.3 &76.1 &73.2 &80.5 \\
    MUTAN ~\citep{ben2017mutan} & 58.1 & 34.1 &73.9 & 76.8 & 73.6 & 81.7 \\
    BAN ~\citep{nguyen2019overcoming,kim2018bilinear}&58.3 &37.4 &72.1 &76.3 &74.6 &79.1 \\
    %MEVF+SAN~\citep{med_nguyen2019overcoming}&64.1&49.2&73.9&76.5&75.3&78.4 \\
    % DAN Fw.~\citep{nam2017dual} &65.4 &49.2 & 76.1&77.2 &74.4&81.5\\
    MEVF+BAN~\citep{nguyen2019overcoming}&66.1 &49.2 &77.2 &78.6 &77.8 &79.8\\ 
    CP+BAN~\citep{liu2021contrastive} & 68.1& 53.1 & 77.9& 80.9&79.1&83.7\\
    CMSA-MTPT~\citep{gong2021cross} & 71.4 & 60.9 & 78.3 & 80.5 & 78.2 & 84.0\\
    MMBERT* \citep{khare2021mmbert} & 72.0 & \bf63.1 & 77.9 & - &- & -\\
    MedViLL* \citep{moon2021multimodal}& 70.9  & 59.7  & 78.2 &  -  & -  &  -  \\
    % \textcolor{blue}{M3AE* \citep{chen2022m3ae}}  & \bf77.0 &  \bf67.2  & \bf83.46 & 83.3 & 80.3 & \bf87.8        \\
    % MEVF+BAN+CR \citep{zhan2020medical} & 71.6& 60.0 & 79.3 &80.0 &78.8 & 82.0\\
    % CP+BAN+CR \citep{liu2021contrastive}& 72.5 & 60.5&80.4 &
    % \textbf{81.9} & \textbf{80.5} & \textbf{84.1}\\
    % \midrule
    % Resnet-101+VLT (w/o pre-training)& 64.1 & 48.6 & 74.3 & 
    % 77.7 & 77.2 & 78.4\\
    % Resnet-101+VLT (w/ pre-training) & 67.5 & 52.9 & 76.3  &
    % 79.8 & 78.3 & 82.2\\
    % Swin-S+VLT (w/o pre-training) & 64.3  & 49.7 & 73.9 & 
    % 80.2 & 78.6 & 82.6 \\
    Swin-S+VLT (w/o pt)  &  66.5  &  53.6 & 75.0  &  80.5 &  79.2 & 82.5\\
    Swin-S+VLT (w/ pt)   & \bf72.1 & 60.9 & \bf79.4 & \bf84.0 & \bf81.9 & \bf87.3 \\ %\textbf{83.9} & \textbf{82.2} & \textbf{86.5}
    % Swin-S+VLT (w/ pre-training)& \bf68.7 & \bf55.3 & \bf77.6 & 
    % \textbf{81.8} & \textbf{80.3} & \textbf{84.1}\\
    
    % Swin-S+VLT (MLM pre-training)& \textbf{75.7} & \textbf{71.1} & 78.2 & 
    % \textbf{81.8} & \textbf{80.3} & \textbf{84.1}\\

    \bottomrule
    \end{tabular}}
    \caption{Test accuracy (\%) of our VLTs and baselines on VQA-RAD and SLAKE. * indicates the results are not reproduced but copied from the original paper due to unavailability of open source codes or key information not provided in the released code. 
    %$\dag$ means no open source code available.
    }
    \label{tab:vqa_result}
\end{table*}

Since both RGC and VQA-RAD are collected from MedPix, they have some overlapping data. However, it does not influence the conclusions drawn from our experiments. This is because the train-test split of VQA-RAD is based on questions instead of images (each image is associated with several questions), and the training set contains all the images in VQA-RAD. So, when a model is trained/fine-tuned on VQA-RAD, it will see all the images anyway. To support this claim, we exclude the 188 images overlapped with VQA-RAD from our RGC dataset and use the rest for pre-training. The results in Table~\ref{tab:overlapping} show that prediction accuracy on the test set of VQA-RAD is 69.80 ± 0.84 (averaged over 30 runs), which is very close to the result of 69.77 ± 0.66 obtained by using all images in RGC for pre-training. This supports our statement that the overlapping images used in pre-training will not affect the conclusions drawn from the reported results.

\begin{table*}[t]
\centering

\scalebox{0.77}{
%\resizebox{width}{!}{

\begin{tabular}{c|c|ccccccc}
\toprule
  Dataset     &  Methods   & BLEU-1 & BLEU-2&  BLEU-3 & BLEU-4 & METEOR & ROUGE-L & CIDER-D \\
\midrule
\multirow{8}*{MIMIC-CXR} 
% & SA\citep{vinyals2015show} & 0.299  & 0.184 & 0.121 & 0.084  & 0.124 & 0.263 & - \\
% & Att2in~\citep{rennie2017self}  & 0.325  & 0.203 & 0.136 & 0.096  & 0.134 & 0.276 & -\\
% & AdaAtt~\citep{lu2017knowing}  & 0.299  & 0.185 & 0.124 & 0.088  & 0.118 & 0.266 & -\\
% & Topdown~\citep{anderson2018bottom} & 0.317  & 0.195 & 0.130 & 0.092  & 0.128 & 0.267 & -\\
% & R2GEN-Base & 0.314 & 0.192 & 0.127 & 0.090 & 0.125 & 0.265 &- \\
& R2GEN~\citep{r2gen}                & 0.353 & 0.218 & 0.145 & 0.103 & 0.142 & 0.277 &- \\

& CMCL*~\citep{liu-etal-2021-competence} & 0.344 & 0.217& 0.140 & 0.097 & 0.133 & 0.281 & -\\

& PPKED*~\citep{liu2021exploring}       & 0.360 & 0.224 & 0.149 & 0.106 & 0.149 & 0.284 & - \\

& KGAE*~\citep{KGAE}                   & 0.369 & \bf0.231 & \bf0.156 & \bf0.118 & 0.153 & 0.295 & - \\ 

& Clinical-BERT*~\citep{yan2022clinical} & \bf0.383 & 0.230 & 0.151 & 0.106 & 0.144 & 0.275 & - \\

% & Resnet101+VLT (w/o pre-train + MT)& 0.354  & 0.206 & 0.128 & 0.084  & \bf0.146 & 0.238 &-\\
% & Resnet101+VLT (w/o pre-train + WST)& 0.317  & 0.189 & 0.121 & 0.083  & 0.130 & \textbf{0.298} &-\\
& Resnet-101+VLT(w/ pt)             & 0.339  & 0.197 & 0.124 & 0.093  & 0.154 & 0.298 &-\\ %0.359  & 0.208 & 0.129 & 0.085  & 0.145 & 0.240 &-
% & Swin-S+VLT (w/ pt + MT) & \textbf{0.362}  & 0.209 & 0.129 & 0.085  & \textbf{0.147} & \textbf{0.289} &-\\
% & Swin-S+VLT (WST + w/ pt) & 0.324  & 0.193 & 0.125 & 0.087  & 0.133 & \textbf{0.300} &-\\
& Swin-S+VLT (w/o pt) & 0.340  & 0.198 & 0.127 & 0.090  & 0.146 & 0.289 & -\\ %?
% & Swin-S+VLT (MT + w/ pt)  & 0.362  & 0.208 & 0.126 & 0.083  & 0.147 & 0.288 &-\\ % rgc pretrain only
& Swin-S+VLT (w/ pt)  & 0.340  & 0.209 & 0.139 & 0.099  & \bf0.166 & \bf0.306 &-\\ % rgc+medicat+roco

\midrule

\multirow{8}*{IU X-Ray} 
& HRGR~\citep{li2018hybrid}  & 0.438 & 0.298 & 0.208 & 0.151 & - & 0.322 & - \\

& CMAS-RL~\citep{jing2020show}  & 0.464 & 0.301 & 0.210 & 0.154 & - & 0.362 & - \\

& R2GEN~\citep{r2gen}               & 0.470    & 0.304  & 0.219  & 0.165 & 0.187 & 0.371  & - \\

& CMCL*~\citep{liu-etal-2021-competence} & 0.473 & 0.305  & 0.217 & 0.164 & 0.186 & 0.378  & - \\

& PPKED*~\citep{liu2021exploring} & 0.483      & 0.315   & 0.224  & 0.168  & 0.190 & 0.376  &  0.351 \\

& KGAE*~\citep{KGAE}          & \bf0.512 &  0.327 & \bf0.240 & \bf0.179 & \bf0.195   & 0.383 & - \\

& Clinical-BERT*~\citep{yan2022clinical}  & 0.495  & \bf0.330   &  0.231 & 0.170 & -   &  0.376  &  0.432 \\

& Resnet-101+VLT (w/ pt)& 0.423 & 0.266&  0.186  & 0.136 & 0.180  & 0.393 & 0.388\\

& Swin-S+VLT (w/o pt) & 0.461 & 0.297  & 0.214 & 0.154  & 0.190 & \bf0.404 & \bf0.462 \\

& Swin-S+VLT (w/ pt) & 0.429 & 0.265  & 0.185 & 0.137  & 0.184  & 0.387 & 0.439 \\

\midrule

\multirow{4}{*}{\DATASETNAMEabbr{}} 
% SA\&T\citep{xu2015show} & 0.296   & 0.227 & 0.197 & 0.181  & 0.141 & 0.254 &  1.315\\
& R2GEN~\citep{r2gen}    & 0.404  & 0.335 & 0.316 & 0.298  & 0.193 & 0.360 & 2.381 \\

& Resnet-101+VLT (w/ pt) &  0.403    & 0.352    & 0.330    & 0.318   & 0.214    & 0.357 & 2.554 \\
& Swin-S+VLT (w/o pt)   & 0.490  & \bf0.453   & 0.435  & 0.419   & 0.280       & \bf0.459 & \bf3.645  \\
& Swin-S+VLT (w/ pt)    & \bf0.491   & 0.452 &  \bf0.436  & \bf0.420  & \bf0.282 & 0.455 & 3.535  \\

\bottomrule
\end{tabular}
}

\caption{Results of generation tasks on MIMIC-CXR, IU X-Ray, and \DATASETNAMEabbr{}. BLEU-N (N=1,2,3,4)~\citep{bleu}, METEOR~\citep{meteor}, ROUGE-L~\citep{rouge}, and CIDER-D~\citep{cider} scores are reported. * indicates the results are not reproduced but copied from the original paper due to unavailability of open source codes.}
\label{table:report_generation_results}
\end{table*}

\begin{table*}[t!]
        \centering
        \scalebox{0.83}{
            \centering
            \begin{tabular}{c|ccc|ccc|ccc|ccc}
            \toprule
            &  \multicolumn{6}{c|}{\DATASETNAMEabbr{}}&\multicolumn{6}{c}{IU X-Ray}\\
            &  \multicolumn{3}{c|}{Text Retrieval}&\multicolumn{3}{c|}{Image Retrieval} & \multicolumn{3}{c|}{Text Retrieval}&\multicolumn{3}{c}{Image Retrieval}\\
             Models   &  R@1 &  R@5 & R@10 & R@1 &  R@5 & R@10 &  R@1 &  R@5 & R@10 & R@1 &  R@5 & R@10   \\
            \midrule
            VSE++ & 26.27 & 43.27 & 50.89 & 30.02 & 47.70 & 55.71     & 0.21 & 3.43 & 5.38  & 0.57 & 3.24 & 5.83 \\
            % Resnet-101+VLT (w/o pre-training)& 7.46 & 19.34 &27.12 
            % & 11.29 & 25.93 & 34.04\\
            Resnet-101+VLT (w/ pt)  & 24.18 & 50.46 &  62.36 & 30.77 & 59.00 & 70.09  & 1.02 & 3.39 &5.59 & 1.19 & 4.24 & 7.12   \\
            Swin-S+VLT (w/o pt) & 17.02 & 38.52 & 51.70 & 18.31 & 43.71 & 57.16  &  0.68 & 2.37  & 4.24 &  0.34 & 1.86 & 5.59 \\
            % Swin-S+VLT (\DATASETNAMEabbr{} only) & 29.15  & 50.49 & 59.94 & 35.41 & 57.56 & 66.91 \\
            Swin-S+VLT (w/ pt) & \bf29.42  & \bf54.59 & \bf67.55 &  \bf34.72 & \bf63.76 & \bf75.66 & \bf1.53 & \bf6.78 & \bf10.51 & \bf2.03 & \bf5.93 & \bf10.00 \\
            \bottomrule
            \end{tabular}
        }
        \caption{Image-text retrieval results on \DATASETNAMEabbr{} and IU X-Ray. Recall@$k$ (\%) is used as the evaluation metric.}
        \label{tab:retrieval_result}
\end{table*}

\subsection{Effect of Pre-training on Med-VL Tasks} \label{sec:downstream_tasks}
In this section, we evaluate the effectiveness of the pre-trained VLT on three downstream medical VL tasks. Unless otherwise stated, the VLT is pre-trained with MLM on a combination of three datasets including \DATASETNAMEabbr{}, ROCO, and MedICaT.

\noindent\textbf{Medical Visual Question Answering.} We conduct experiments on VQA-RAD and SLAKE and compare VLT with general VQA models (\ie, SAN~\citep{yang2016stacked}, MFH~\citep{yu2018beyond}, MCB~\citep{fukui2016multimodal}, MUTAN~\citep{ben2017mutan}, BAN~\citep{kim2018bilinear}) as well as Med-VQA models (\ie, MEVF~\citep{nguyen2019overcoming}, CPRD~\citep{liu2021contrastive}, CMSA-MTPT~\citep{gong2021cross}, MMBERT~\citep{khare2021mmbert}, MedViLL~\citep{moon2021multimodal}, M3AE~\citep{chen2022m3ae}). 
Specifically, MEVF is trained in a semi-supervised manner to overcome the lack of medical training data. Both CPRD and CMSA-MTPT propose to pre-train three visual feature extractors (\ie, Resnet) for three image modalities (\ie, abdomen CT, chest X-ray and brain MRI) respectively, on current datasets~\citep{vqa-rad,slake}, which can be integrated with cross-modal networks such as BAN and BERT~\citep{devlin2018bert}. CR utilizes question types to improve model performance. MMBERT and MedViLL pre-train a VLT on ROCO and MIMIC-CXR, respectively. %\textcolor{blue}{M3AE~\citep{chen2022m3ae} utilizes MAE~\citep{MAE} to further improve the pre-training performance.}
%are similar to our model with less pre-training data, but their codes don't provide key information to recover the results.
To have a fair comparison, we do not compare with previous works that use additional label information as in~\citet{zhan2020medical, liu2022medical} or external knowledge as in ~\citet{MM2022-VLP}. Following most previous works, we report test accuracy on VQA-RAD and SLAKE.
We can observe from the results in \Cref{tab:vqa_result} that:
\begin{itemize}[leftmargin=5mm]
    \item Pre-training can significantly improve model performance on Med-VQA tasks.
    \item Our pre-trained VLT outperforms state-of-the-art methods on two benchmarks.
\end{itemize}

\noindent\textbf{Medical Report Generation.} 
We conduct experiments on MIMI-CXR, IU X-Ray~\citep{iuxray}, and \DATASETNAMEabbr{}. 
Current works on report generation~\citep{jing2017automatic, li2018hybrid, li2019knowledge, jing2020show, r2gen, liu-etal-2021-competence, r2gen, liu2021exploring, yan2022clinical, KGAE, jain2021radgraph} mainly focus on chest X-ray images and experiment with MIMIC-CXR and IU X-Ray. 
% \citep{jing2017automatic} proposes a hierarchical LSTM based on the co-attention mechanism to generate reports. \citep{li2018hybrid} leverages pre-defined templates to tackle this task. 
R2GEN~\citep{r2gen} proposes a relational memory and memory-driven layer normalization in a standard Transformer for generation. CMCL~\citep{liu-etal-2021-competence} proposes to use the most suitable samples to train the model based on current model competence. 
PPKED~\citep{liu2021exploring} proposes a knowledge explorer and distiller to generate reports. KGAE~\citep{KGAE} leverages an external knowledge graph.
We follow R2GEN to pre-process MIMI-CXR and IU X-Ray and compare with existing works using the same pre-processing pipeline. Beam search with the beam size 3 is used. CMCL, PPKED, KGAE, and Clinical-BERT are tailored for generating reports for chest X-ray images, and the source codes are not released. Hence, we only compare with R2GEN on \DATASETNAMEabbr{}. 
% \textcolor{red}{Following previous generation works, we report BLEU-N (N=1,2,3,4)~\citep{bleu}, METEOR~\citep{meteor}, ROUGE-L~\citep{rouge}, and CIDER-D~\citep{cider} scores of the generated reports.} 
The results in \Cref{table:report_generation_results} suggest:
%and we can make the following observations:
\begin{itemize}[leftmargin=5mm]
    \item Pre-training does not help to improve the performance on generation tasks and can even decrease the model performance on IU X-Ray. It is probably because generation tasks are different and more complicated than classification tasks (\eg, Med-VQA and image-text retrieval), which suggests the inadequacy of current pre-training data and strategy for generation tasks. 
    %We also suspect that although all datasets used for training are on radiology, there is still a gap between these datasets in both images and vocabularies, making pre-training ineffective. % This may imply that the pre-training of generative models requires much more data like GPT~\citep{gpt3}, and current medical datasets are still not large enough.
    \item
    Resnet and Swin Transformer achieve comparable performance on MIMIC-CXR and IU X-Ray, but there is a large gap between their performance on \DATASETNAMEabbr{}. It shows that \DATASETNAMEabbr{} as a multi-modality dataset is a better benchmark for report generation than single-modality datasets, as it can reflect the difference in model capacity for learning visual features.
    %can enlarge the performance gaps between visual backbones, while in MIMIC-CXR and IU X-Ray, Resnet and Swin Transformer achieve similar scores. We guess datasets with more imaging modalities are more demanding on visual modules.
\end{itemize}

\noindent\textbf{Medical Image-text Retrieval.} We conduct experiments on \DATASETNAMEabbr{} and IU X-Ray. 
% Only images of AP and PA views from MIMIC-CXR are used. 
Previous works on this task~\citep{hsu2018unsupervised, moon2021multimodal, wang2021self, huang2021gloria, zhang2020contrastive} are mainly designed for chest X-ray datasets, \eg, MIMIC-CXR and CheXpert ~\citep{irvin2019chexpert}. However, CheXpert is not released, and the source codes for \citep{zhang2020contrastive, wang2021self, hsu2018unsupervised} are unavailable. 
% And recent image-text retrieval models such as VSRN~\citep{li2019vsrn} and SCAN~\citep{lee2018stacked} heavily rely on bounding box features from Faster R-CNN. %, which make them hard to be applied in the medical domain. 
Hence, we only compare with VSE++~\citep{faghri2018vse}, a strong baseline for image-text retrieval. %that utilizes Resnet as the visual extractor.
Following previous works, we report Recall@$K$ ($K=1,5,10$). The results presented in \Cref{tab:retrieval_result} suggest that:
\begin{itemize}[leftmargin=5.5mm]
    \item Pre-training can greatly improve model performance on image-text retrieval tasks.
    %although the downstream datasets are not included in the pre-training datasets.
    \item Swin Transformer significantly outperforms Resnet on retrieval tasks, just as in Med-VQA tasks and report generation tasks on \DATASETNAMEabbr{}.
    %pre-training can still help improve Although pre-training data doesn't contain IU X-ray, 
    %\item Similar to the observation on report generation,  \DATASETNAMEabbr{} as a multi-modality dataset is a better benchmark for image-text retrieval than single-modality datasets such as MIMIC-CXR, as it can reflect the difference in capacity.
\end{itemize}

\section{Conclusions and Limitations}
\label{sec:conclusion}

%The main contributions of this work are also two-fold: 
% \begin{itemize}
% \item 
This paper makes two main contributions. 1) We present a comprehensive empirical study on medical VLP, providing analysis on pre-training decisions and evaluating the effectiveness of the pre-trained VLTs on both generation and understanding Med-VL tasks. We distill the experimental results into several key observations which can be used as a guide to future VLP research. The pre-trained VLTs can also serve as strong baselines for future research. 2) We propose RGC, a high-quality radiographic VL dataset of multiple imaging modalities, which can be used as a pre-training dataset or a new benchmark for medical report generation and medical image-text retrieval, to supplement the very small pool of existing Med-VL benchmarks. One limitation of RGC is that its size is relatively small. However, since the MedPix database keeps growing, it provides opportunities to expand and improve RGC in the future.

\section*{Acknowledgment}
We sincerely thank the MedPix team including Dr. Dina Demner-Fushman, Mr. Soumya Gayen, and Dr. James G. Smirniotopoulos for their kind help in hosting the RGC dataset on MedPix website. We would also like to thank the anonymous reviewers for their helpful comments. This research was partially supported by the grant of project P0038194 (1-ZVVX) funded by PolyU (UGC).

\bibliography{jmlr-sample}

\clearpage

\appendix

\section{Additional Experiments}
\label{sec:appendix}

% \textbf{SLAKE} \textcolor{red}{contains 140 head CT/MRI images, 41 neck CT images, 219 chest X-Ray/CT images, 201 abdomen CT/MRI images, and 41 pelvic cavity CT images. In total, there are 282 CT images, 181 MRI images, and 179 X-Ray images.}

% \textbf{VQA-RAD} \textcolor{red}{contains 104 head axial single-slice
% CT/MRI images, 107 chest X-Ray images, and 104 abdominal axial CT images.}

\subsection{More Ablation Studies on Pre-training Settings}
We provide additional experimental results of VLTs pretrained with different visual backbones and different datasets on downstream tasks including Med-VQA, report generation, and medical image-text retrieval.

\textbf{Med-VQA.}  The results are shown in \Cref{tab:backbone_data_med_vqa2}, and we can make the following observations:

\begin{itemize}[leftmargin=5.5mm]
    %\item 
    %Large-scale data can help improve pre-trained VLTs with different backbones on both Med-VQA benchmarks.
    
    \item For each different visual backbone, the VLT pre-trained with larger dataset (\ie, \DATASETNAMEabbr{} + ROCO + MedICaT) consistently achieves better performance on both Med-VQA benchmarks. 

    \item When pre-training with \DATASETNAMEabbr{} + ROCO + MedICaT, ResNet and ViT achieve similar performance as Swin Transformer on VQA-RAD , but the latter performs much better on SLAKE.
\end{itemize}

\textbf{Report Generation.} The results are shown in \Cref{tab:data_visual_report_generation_results2}. We can observe:

\begin{itemize}[leftmargin=5.5mm]
    \item Pre-training with larger dataset (\ie, \DATASETNAMEabbr{} + ROCO + MedICaT) does not help to improve model performance on downstream generation tasks. 
    %Using large-scale datasets for pre-training doesn't necessarily benefit the performance of VLTs with different backbones on generation tasks. 
    
    \item The VLTs with Swin-S as visual backbone achieve the best performance, though pre-training does not seem to make a difference.
    %The VLT with Swin-S achieves similar results with the one pre-trained by \DATASETNAMEabbr{} only. 
    %The possible reason is that we adopt the bidirectional MLM and seq2seq MLM as the pre-training objectives, while we fine-tune the model with the seq2seq MLM. Hence, on report generation task of \DATASETNAMEabbr{}, the only difference between the model trained from scratch and the one pre-trained by \DATASETNAMEabbr{} is the bidirectional MLM which does not help improve the model performance on generation tasks.
\end{itemize}

\textbf{Medical Image-text Retrieval.}  The results are summarized in \Cref{tab:retrieval_result_add2}, and we can make the following observations:

\begin{itemize}[leftmargin=5.5mm]
    \item Pre-training is highly effective for downstream medical image-text retrieval tasks. For each visual backbone, more pre-training data leads to overall better performance.
    
   % \item Although VLT is pre-trained without \DATASETNAMEabbr{}, pre-training can still improve the model performance on medical image-text retrieval to a great extent. 
    % \item The models jointly pre-trained by \DATASETNAMEabbr{}, ROCO and MedICaT can outperform the one pre-trained by \DATASETNAMEabbr{} in R@5 and R@10. This shows using large-scale data for pre-training can improve the generalization ability of the models.
    
    \item The VLTs with Swin-S as visual backbone outperform others.
    %Swin Transformer outperforms other visual encoders.

\end{itemize}

\subsection{Pre-training with Out-of-domain Datasets}
To demonstrate the importance of using in-domain data for pre-training, we use MSOCO Captions~\citep{chen2015mscococaption} and Conceptual Captions (CC)~\citep{sharma2018conceptual} to pre-train VLTs with Swin-S as visual backbone and report the results on Med-VQA tasks in \Cref{tab:out_of_domain_data__med_vqa2}. We can observe that:

\begin{itemize}[leftmargin=5.5mm]

\item Pre-training with a small in-domain dataset (\eg, RGC) is much more effective than with large-scale out-of-domain datasets.

\item Pre-training with out-of-domain datasets may have an adverse effect and significantly decrease model performance.
    %\item Large-scale out-of-domain pre-training datasets are less effective than in-domain datasets on Med-VQA.
    
    %\item  Low-quality and out-of-domain datasets can even decrease the model performance.
\end{itemize}

% \section{Distribution of Imaging Modalities in RGC}
% Our proposed RGC dataset contains radiographic images of different modalities, as summarized in  
% \Cref{tab:rgc_modality}.
% %numbers of images of different modalities on \DATASETNAMEabbr{}. 

% \subsection{Impact of the Overlapping Images between RGC and VQA-RAD on Pre-training Performance}\label{app:overlap}
% We exclude the 188 images overlapped with VQA-RAD from our RGC dataset and use the rest for pre-training. The results in Table~\ref{tab:overlapping} show that prediction accuracy on the test set of VQA-RAD is 69.80 ± 0.84 (averaged over 30 runs), which is very close to the result of 69.77 ± 0.66 obtained by using all images in RGC for pre-training. This supports our statement that the overlapping images used in pre-training will not affect the conclusions drawn from the reported results.

\begin{table}[t!]
     \centering
    \scalebox{0.8}{
    \begin{tabular}{l|c|c|c}
    \toprule
     Datasets & Samples & VQA-RAD &  SLAKE  \\
    \midrule
     None           &  N/A & 65.28 $\pm$ 0.98 &  79.74 $\pm$ 0.97 \\
    \DATASETNAMEabbr{} & $\sim$17k  &  69.77 $\pm$ 0.66 &  81.50 $\pm$ 0.36\\
    
    MSCOCO           &  $\sim$414k  &  62.86 $\pm$ 1.67 &  80.57 $\pm$ 0.32  \\ % lr  VQA-RAD
    CC               &  $\sim$2M    &  59.31 $\pm$ 1.32 &  79.23 $\pm$ 0.48  \\ % 150w  ?
    % MSCOCO+CC        &  $\sim$2.4M  & 64.08 $\pm$ 1.04 &  80.28 $\pm$ 0.47   \\ 
    \bottomrule
    \end{tabular}
    }
    \caption{Comparisons of VLTs pre-trained with in-domain and out-of-domain datasets. The visual backbone is Swin-S.} % Note the numbers for pre-training may differ from the original datasets as we filter out some samples.
\label{tab:out_of_domain_data__med_vqa2}
\end{table}

\begin{table*}[ht!]
    \centering

    %\begin{tabular}{m{3.2cm}|c|c}
    \begin{tabular}{c|ccc|c|c}
    \toprule
    Visual Backbone &   \DATASETNAMEabbr{} & ROCO & MedICaT & VQA-RAD &  SLAKE  \\
    \midrule
    Linear/16   &\cmark & \xmark  & \xmark   & 69.56 $\pm$ 0.96 &  78.23 $\pm$ 0.36 \\
    Linear/16   &\cmark & \cmark  & \cmark   & 70.41 $\pm$ 0.41 &  79.76 $\pm$ 0.57 \\ 
    ViT-B/16    &\cmark & \xmark  & \xmark  & 67.81 $\pm$ 1.07 &  79.25 $\pm$ 0.48 \\
    ViT-B/16    &\cmark & \cmark  & \cmark   & \bf71.21 $\boldsymbol\pm$ 0.89 &  80.96 $\pm$ 0.63 \\ 
    Resnet-101  &\cmark & \xmark  & \xmark   &  69.62 $\pm$ 1.03 &  80.03 $\pm$ 0.49\\
    Resnet-101  &\cmark & \cmark  & \cmark   & \bf71.27 $\pm$ 0.46 &  81.39 $\pm$ 0.64\\ 
    Swin-S      &\cmark & \xmark  & \xmark   & 69.77 $\pm$ 0.66 & 81.50 $\pm$ 0.36\\
    Swin-S      &\cmark & \cmark  & \cmark   & \bf71.20 $\boldsymbol\pm$ 0.83 &  \bf83.40 $\boldsymbol\pm$ 0.57\\ 
    \bottomrule
    \end{tabular}
        \caption{Comparison of VLTs pre-trained with different visual modules and different datasets for Med-VQA.}
    \label{tab:backbone_data_med_vqa2}
\end{table*}

\begin{table*}[ht!]
\centering
\scalebox{0.75}{
%\resizebox{width}{!}{
    \begin{tabular}{c|ccc|ccccccc}
    \toprule
    Visual Backbone   &   \DATASETNAMEabbr{} & ROCO & MedICaT   & BLEU-1 & BLEU-2&  BLEU-3 & BLEU-4 & METEOR & ROUGE-L & CIDER-D \\
    \midrule
    
        Linear/16   &\cmark & \xmark  & \xmark   & 0.332 & 0.285 & 0.264 & 0.252 & 0.195 & 0.342 & 2.291\\ % ok
        Linear/16   &\cmark & \cmark  & \cmark   & 0.346 & 0.302 & 0.281 & 0.270 & 0.213 & 0.361 & 2.500 \\ % ok
        ViT-B/16    &\cmark & \xmark  & \xmark  &  0.417 & 0.373 & 0.351 & 0.339 & 0.244 & 0.408 & 3.066 \\ % ok
        ViT-B/16    &\cmark & \cmark  & \cmark   & 0.404 & 0.365 & 0.338 & 0.326 & 0.241 & 0.397 & 2.919 \\  % ?
        Resnet-101  &\cmark & \xmark  & \xmark   &  0.404 & 0.359  & 0.332  & 0.320 & 0.218  & 0.361 & 2.560 \\ % 
        Resnet-101  &\cmark & \cmark  & \cmark   &  0.403  & 0.352 & 0.330  & 0.318 & 0.214  & 0.357 & 2.554 \\ % 
        Swin-S     &\xmark & \xmark  & \xmark  & 0.490  & 0.453 & 0.435 & 0.419 & 0.280  & \bf0.459 & 3.645  \\ %
        Swin-S     &\cmark & \xmark  & \xmark  & 0.491  & \bf0.455 & 0.435 & 0.420 & 0.281  & 0.455 & \bf3.676  \\ % 
        Swin-S     &\cmark & \cmark  & \cmark  & \bf0.491  & 0.452 &  \bf0.436  & \bf0.420  & \bf0.282 & 0.455 & 3.535  \\ %
    
    \bottomrule
    \end{tabular}
}
\caption{Comparison of VLTs pre-trained with different visual modules and different datasets for report generation on \DATASETNAMEabbr{}.}
\label{tab:data_visual_report_generation_results2}
\end{table*}

\begin{table*}[ht!]
    \centering
\scalebox{0.9}{
    \begin{tabular}{c|ccc|ccc|ccc}
    \toprule
    % \hline
    % \hline
    \multirow{2}*{Visual Backbone} &  \multicolumn{3}{c|}{Pre-training datasets} &  \multicolumn{3}{c|}{Text Retrieval}&\multicolumn{3}{c}{Image Retrieval}\\
     & \DATASETNAMEabbr{} & ROCO &  MedICaT   &  R@1 &  R@5 & R@10 & R@1 &  R@5 & R@10  \\
    \midrule
    % VSE++ & 27.77 & 45.06 & 54.02 & 31.98 & 50.14 & 58.73\\
    Linear/16 &   \xmark & \xmark & \xmark & 15.31 & 26.39 & 34.72 & 15.52 & 31.96 & 40.08 \\
    Linear/16  & \cmark & \xmark & \xmark &  18.12 & 28.07 & 32.72 & 24.50 & 34.45 & 39.81  \\
    Linear/16 & \cmark & \cmark & \cmark  &  21.85 & 40.78   & 51.27 &  27.37 & 48.40 & 57.54  \\
    
    ViT-B/16  &   \xmark & \xmark & \xmark & 3.19 & 9.30 & 14.44 & 4.65 & 12.93 & 19.69 \\
    ViT-B/16  & \cmark & \xmark & \xmark & 19.04 & 37.21 & 45.43 & 26.07 & 47.97 & 54.89   \\
    ViT-B/16 & \cmark & \cmark & \cmark  & 17.39 & 39.16 &  52.24 & 22.96 & 47.81 & 61.17 \\
    Resnet-101 & \xmark & \xmark & \xmark & 5.30 & 12.49 & 18.71 & 7.41 & 18.17 & 26.01 \\% 7.46 & 19.34 &27.12  & 11.29 & 25.93 & 34.04\\
    Resnet-101 & \cmark & \xmark & \xmark & 22.71 & 39.05 & 46.46 & 31.10 & 46.78 & 54.30\\ % 25.43 & 43.17 &  53.05 & 31.93 & 51.97 & 61.53 \\
    Resnet-101 & \cmark & \cmark & \cmark  & 24.18 & 50.46 &  62.36 & 30.77 & 59.00 & 70.09  \\
    Swin-S  &   \xmark & \xmark & \xmark & 17.02 & 38.52 & 51.70 & 18.31 & 43.71 & 57.16\\
    Swin-S  & \cmark & \xmark & \xmark  & \bf30.86 & 46.22 & 54.68 & \bf35.32 & 56.30 & 63.17 \\ % 29.15  & 50.49 & 59.94 & 35.41 & 57.56 & 66.91 \\
    Swin-S  & \xmark & \cmark & \cmark  & 22.88 & 47.65 & 61.71 & 29.53 & 57.17 & 71.61 \\ 
    Swin-S  & \cmark & \cmark & \cmark  & 29.42  & \bf54.59 & \bf67.55 &  34.72 & \bf63.76 & \bf75.66 \\
    \bottomrule
    \end{tabular}
}
    \caption{Comparison of VLTs pre-trained with different visual modules and different datasets for medical image-text retrieval on \DATASETNAMEabbr{}.}
    \label{tab:retrieval_result_add2}
\end{table*}

% \subsection{Statistics of Med-VQA Datasets}
% \label{appendix:stat_vqa}

\end{document}